\documentclass[sigconf]{acmart}
\makeatletter
\@ifundefined{correspondingauthor}{%
\newcommand\correspondingauthor{%
  \if@ACM@anonymous\else
  \g@addto@macro\addresses{\@correspondingauthormark}%
  \fi}
\def\@correspondingauthormark{\g@addto@macro\@currentauthors{%
    \advance\hfuzz by 5pt\relax\textsuperscript{\ding{41}}\relax}}
}{}
\makeatother

\usepackage{multirow}
\usepackage[ruled,vlined,linesnumbered]{algorithm2e}
\usepackage{bm}
\usepackage{adjustbox}
\usepackage{graphicx} 
\usepackage{caption}
\usepackage{booktabs}
\usepackage{tabularx}
\usepackage{array}
\usepackage{subcaption}
\usepackage{color}
\usepackage{listings}
\usepackage{amsmath}
\usepackage{bbding}
\usepackage{pifont}
\usepackage{wasysym}
\usepackage{colortbl}
\definecolor{lightgray}{RGB}{219,219,219} 
\AtBeginDocument{%
  }

\setcopyright{acmlicensed}
\copyrightyear{2026}
\acmYear{2026}
\setcopyright{cc}
\setcctype{by}

\acmConference[MM '26]{Proceedings of the 34th ACM International Conference on Multimedia}{November 10--14, 2026}{Rio de Janeiro, Brazil}
\acmBooktitle{Proceedings of the 34th ACM International Conference on Multimedia (MM '26), November 10--14, 2026, Rio de Janeiro, Brazil}
\acmDOI{10.1145/3767308.3835115}
\acmISBN{979-8-4007-2213-4/2026/11}

\let\originalPublicationFootnote\footnotetextcopyrightpermission
\renewcommand{\footnotetextcopyrightpermission}[1]{%
  \originalPublicationFootnote{\raggedright
    Proceedings of the 34th ACM International Conference on Multimedia
    (ACM MM~'26), November 10--14, 2026, Rio de Janeiro, Brazil.\par}}

\begin{document}

\title{Disentangled Embedding Optimization for Diffusion-Based Dataset Distillation}

\author{Hongcheng Li}
\orcid{0009-0005-7847-3806}
\affiliation{%
  \institution{Institute of Information Engineering,
Chinese Academy of Sciences \&
School of Cyber Security, University
of Chinese Academy of Sciences \& State Key Laboratory of Cyberspace Security Defense}
  \city{Beijing}
  \country{China}}
\email{lihongcheng@iie.ac.cn}

\author{Yucan Zhou}
\correspondingauthor
\orcid{0000-0002-1316-9118}
\affiliation{%
  \institution{College of Intelligence and Computing,
 Tianjin University}
  \city{Tianjin}
  \country{China}}
\email{zhouyucan@tju.edu.cn}

\author{Yibin Wang}
\orcid{0009-0008-5585-9791}
\affiliation{%
  \institution{Institute of Information Engineering,
Chinese Academy of Sciences \&
School of Cyber Security, University
of Chinese Academy of Sciences \& State Key Laboratory of Cyberspace Security Defense}
  \city{Beijing}
  \country{China}}
\email{wangyibin18@mails.ucas.ac.cn}

\author{Xiaoyan Gu}
\orcid{0000-0003-0673-0058}
\affiliation{%
  \institution{Institute of Information Engineering,
Chinese Academy of Sciences \&
School of Cyber Security, University
of Chinese Academy of Sciences \& State Key Laboratory of Cyberspace Security Defense}
  \city{Beijing}
  \country{China}}
\correspondingauthor
\email{guxiaoyan@iie.ac.cn}

\author{Bo Li}
\orcid{0000-0001-6709-0942}
\affiliation{%
  \institution{Institute of Information Engineering,
Chinese Academy of Sciences \\ State Key Laboratory of Cyberspace Security Defense}
  \city{Beijing}
  \country{China}}
\email{libo@iie.ac.cn}

\author{Weiping Wang}
\orcid{0000-0002-8618-4992}
\affiliation{%
  \institution{Institute of Information Engineering,
Chinese Academy of Sciences }
  \city{Beijing}
  \country{China}}
\email{wangweiping@iie.ac.cn}
\renewcommand{\shortauthors}{Hongcheng Li et al.}

\begin{abstract}
Dataset distillation aims to compress large-scale datasets into compact synthetic subsets that enable efficient model training with comparable performance to the original data. Recent advances show that diffusion models are highly effective for this task due to their strong generative capability. Existing diffusion-based approaches typically perform class-wise clustering and use the obtained cluster prototypes as diffusion inputs to generate distilled samples. 
However, the uniformly distributed original samples in a category inevitably lead to entanglement near cluster boundaries, pulling the cluster prototypes closer and reducing the inter-cluster separation. Consequently, the diffusion model receives overly similar latent inputs, resulting in synthetic data with limited diversity.
To address this issue, we propose Disentangled Embedding Optimization (DEO), a simple yet effective framework that replaces static clustering with trainable embeddings. These embeddings are optimized to match the Gaussian statistics (i.e., mean and covariance) of the original data distribution, while a diversity regularizer is designed to encourage separation among these embeddings to improve the intra-class variability.
Extensive experiments demonstrate that our proposed DEO can generate more diversified and representative distilled data than existing cluster-based diffusion methods, achieving state-of-the-art performance and superior generalization across multiple datasets.  
\end{abstract}



\maketitle

\section{Introduction}

\begin{figure}[t]
  \centering
   \resizebox{1.0\linewidth}{!}{
  \includegraphics[width=1.0\textwidth]{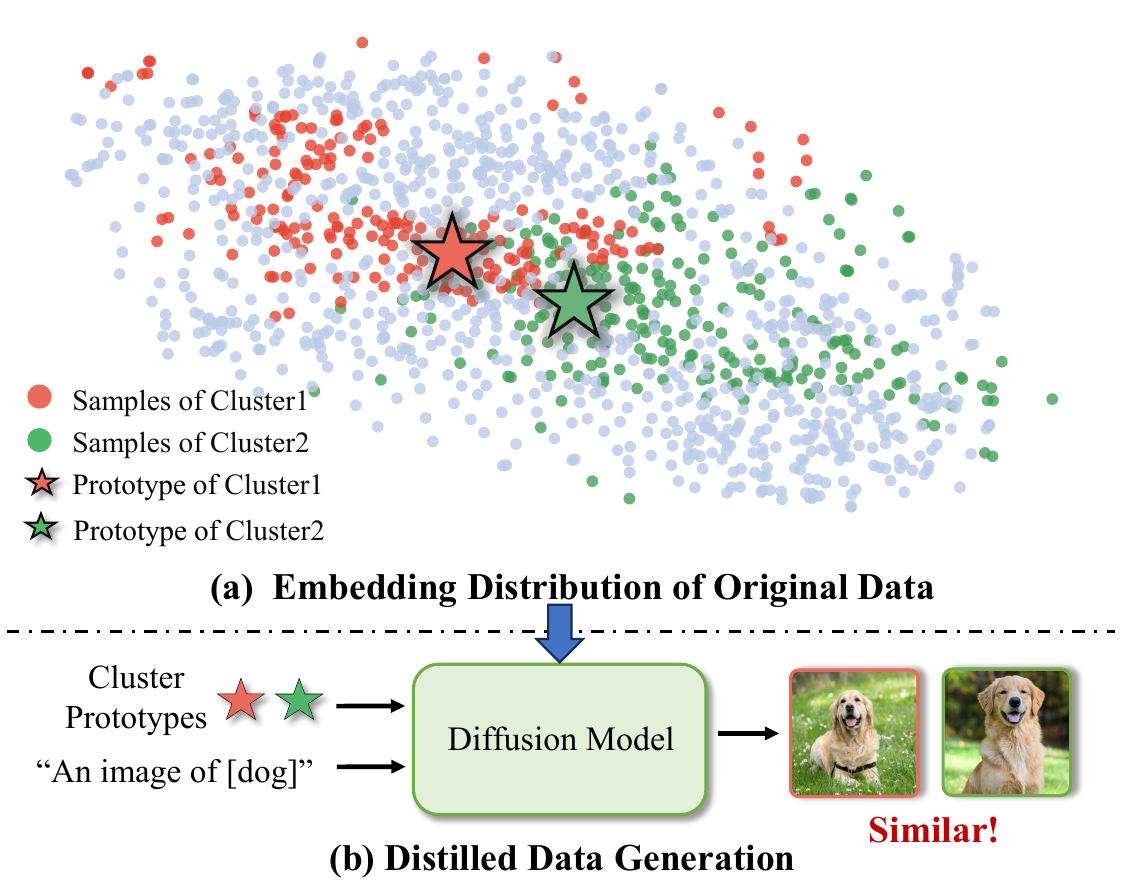}}
\caption{Illustration of the entanglement issue in distilled data generation. (a) t-SNE visualization shows that entangled sub-cluster prototypes in the original data reduce separability. (b) These less-separable latent inputs for the diffusion model result in indistinguishable synthesized images.}

  \label{fig:latent_distribution}
\end{figure}

The success of deep learning in vision, language, and multimodal tasks largely stems from the availability of large-scale datasets~\cite{lin2014microsoft, kuznetsova2020open, 10.1145/3503161.3549202,10.1145/3774904.3792350,11329188}.
However, the growing size of such datasets leads to substantial costs in storage, training time, and computation.
To mitigate these issues, dataset distillation has emerged as a promising paradigm that synthesizes compact datasets capable of retaining the training efficacy of the original datasets~\cite{wang2018dataset, yu2023review,lei2023survey}.
Beyond efficiency, distilled datasets also benefit privacy-sensitive~\cite{dong2022privacy,zheng2025dosser} and resource-constrained scenarios~\cite{yang2023efficient,gu2024ssd}.

Early dataset distillation (DD) methods typically employ a bi-level optimization framework, where the outer loop updates model parameters and the inner loop refines synthetic data by aligning training trajectories \cite{cazenavette2022dataset, du2023minimizing, liu2024att} or gradients \cite{zhao2021datasetcondensation, zhao2021differentiatble, lee2022dataset}. Despite their effectiveness, the interdependent optimization of models and distilled data incurs prohibitive computational costs, often restricting their application to low-resolution datasets like CIFAR-10/100. To mitigate this, subsequent research has decoupled the process by directly matching feature statistics \cite{zhao2023distribution, wang2022cafe, zhao2023improved, yin2023squeeze}. However, these methods still rely on pixel-space optimization, which makes scaling to high-resolution datasets inherently difficult \cite{zhao2025d3hr, chen2025igd, liu2025survey}.

To overcome the limitations of pixel-space optimization, several pioneering studies have incorporated generative priors into dataset distillation. For example, GLAD~\cite{cazenavette2023generalizing} improves distillation by projecting optimization losses onto the intermediate latent layers of a pre-trained GAN, thereby exploiting the structural and semantic information captured at different depths to regularize synthesized samples toward the natural image manifold. However, GAN-based frameworks still require backpropagation through complex generator architectures, resulting in substantial memory overhead.
To alleviate this issue, more recent research has shifted toward diffusion-based distillation. Compared with GANs, diffusion models operate in a highly expressive and semantically linearized latent space induced by a pre-trained VAE. Within this paradigm, the VAE encoder first projects original images into a compact latent manifold, allowing distillation to be performed in a lower-dimensional yet semantically structured space. Specifically,  D$^4$M~\cite{Su_2024_CVPR} facilitates distillation by using $k$-means cluster prototypes as the latent foundation for diffusion. Building upon this paradigm, VLCP~\cite{zou2025vlcp} integrates textual guidance by extracting LLaVA-generated descriptions for each cluster, leveraging these rich semantic prompts to further steer the synthesis of distilled samples.

However, a critical limitation is overlooked: in the latent space of a pretrained VAE encoder, features within each class tend to be uniformly distributed.
Under such conditions, samples lying near the boundary of two clusters are inevitably and often arbitrarily assigned to one cluster or the other, which will reduce the distance between the corresponding cluster prototypes. 
As illustrated in Figure~\ref{fig:latent_distribution}(a), we visualize the latent embeddings of a  class, where samples belonging to two clusters are highlighted in orange and green, respectively.
It is obvious that the entangled samples tend to pull their corresponding cluster prototypes closer in the feature space, thereby reducing inter-cluster separation.
As shown in Figure~\ref{fig:latent_distribution}(b), when these less separable cluster prototypes are used as latent inputs to the diffusion model, the generated distilled images appear visually similar and lack diversity.
Consequently, the synthesized data fail to capture the full spectrum of intra-class semantic variations, limiting the performance of downstream tasks.

To address this issue, we propose Disentangled Embedding Optimization (DEO), a simple yet effective framework for diffusion-based dataset distillation.
Instead of performing clustering analysis in the latent space, our method directly optimizes the latent embeddings, which serve as inputs of the diffusion model to generate distilled data.
Specifically,  we treat the latent embeddings as learnable parameters that are optimized to approximate the original data distribution.
For each class, we estimate a Gaussian distribution in the latent space by computing the empirical mean and covariance matrix with the original data.
Then, the mean and covariance of the trainable latent embeddings are aligned with those of the corresponding class distributions, ensuring semantic representativeness.
To further enhance intra-class diversity of the distilled data, we introduce an explicit dispersion regularization that promotes mutual separation among the trainable latent embeddings.
Finally, the optimized embeddings, together with their corresponding class textual prompts, are fed into the diffusion model to generate  distilled data.
Extensive experiments on CIFAR-10/100, ImageNet-1K, and several ImageNet subsets (ImageWoof, ImageNette, and ImageIDC) demonstrate that our DEO can generate diversified distilled data, which will consistently improve the performance of downstream tasks, all without introducing any external knowledge. 

Our main contributions are summarized as follows:
\begin{itemize}
\item We propose a simple yet effective framework, termed Disentangled Embedding Optimization (DEO) for diffusion-based dataset distillation, which makes the input latent space trainable rather than relying on conventional clustering.

\item  We design a dispersion regularizer that explicitly encourages diversity among synthetic embeddings, effectively mitigating prototype collapse and promoting richer intra-class representations.

\item Extensive experiments on CIFAR-10/100, ImageNet-1K, and multiple ImageNet subsets show that DEO consistently outperforms prior diffusion-based baselines with strong cross-architecture generalization and lower computational cost.
\end{itemize}

\section{Related Work}

\subsection{Matching-Based Methods}

Early approaches to dataset distillation typically adopt a {bi-level optimization framework}, where the outer loop updates model parameters and the inner loop optimizes the distilled data to match either gradient signals or training trajectories. Specially, {gradient matching} aligns the stepwise gradients computed on the original and distilled data~\cite{zhao2021datasetcondensation,zhao2021differentiatble,kim2022dataset,lee2022dataset}, whereas {trajectory matching} extends this idea by aligning the entire optimization paths~\cite{cazenavette2022dataset,cui2023scaling,du2023minimizing,liu2024att}. Despite demonstrating strong performance, these methods are computationally expensive due to the repeated alternation between model updates and data optimization.

To improve efficiency, recent methods leverage {distribution matching}, which aligns feature distributions without updating the  model~\cite{zhao2023distribution}. Early variants typically utilize randomly initialized networks to extract features. For example, CAFE~\cite{wang2022cafe} performs multi-layer prototype alignment; DataDAM~\cite{sajedi2023datadam} selects informative regions via attention mechanisms; and M3D~\cite{zhang2024m3d} conducts feature alignment in a reproducing kernel Hilbert space. Building on this line of work, IDM~\cite{zhao2023improved} and DANCE~\cite{zhang2024dance} enhance representational capacity by employing multiple pretrained extractors, while IID~\cite{deng2024iid} and DSDM~\cite{li2024dsdm} introduce feature moment matching to promote semantic diversity. However, these methods require both the dataset and the model to reside in memory, which limits scalability to large-scale settings.

To address the scalability challenge, decoupled distillation strategies have emerged~\cite{yin2023dataset,sun2024diversity,shao2024gvbsm}. A representative example is SRe2L~\cite{yin2023squeeze}, which first trains a reference model to absorb the knowledge from the full dataset, and then optimizes the distilled data by matching per-batch BatchNorm statistics to the model’s global statistics. Additionally, soft labels generated by the reference model provide supervisory signals for distillation. While this strategy substantially reduces computational cost, it may compromise diversity in synthetic samples, limiting generalization to unseen architectures.
To mitigate this issue, DWA~\cite{dwa2024neurips} introduces training perturbations and variance-aware weighting to encourage diverse sample generation. Complementarily, DEDA~\cite{li2025deda} incorporates a decorrelation regularization term that reduces redundancy among distilled features, thereby further enhancing semantic diversity.
Although the decoupled distillation methods are more computationally efficient, they still follow the pixel-level optimization paradigm, where memory usage and computational cost grow greatly with image resolution, ultimately limiting their scalability to high-resolution datasets.

\subsection{Generative-Based Methods}
Recent research has shifted from pixel-space optimization to leveraging generative priors. As a pioneering work, GLAD~\cite{cazenavette2023generalizing} introduces a deep generative prior from GANs to regularize the distillation process, significantly enhancing cross-architecture generalization. However, GAN-based methods often struggle with high-resolution scaling or necessitate complex, memory-intensive optimization.
Building on the success of generative modeling \cite{liu2025autoprompt}, diffusion models have emerged as robust backbones for dataset distillation due to their superior capacity for high-fidelity image synthesis. For instance, Gu et al.~\cite{gu2024efficient} enhance the representativeness of synthetic data by fine-tuning diffusion models to align predicted embeddings with those of original images. Although effective, this multi-stage fine-tuning pipeline is computationally expensive, especially for large-scale datasets.
To mitigate the overhead, subsequent studies exploit a frozen VAE and perform distillation directly in its latent space. D$^4$M~\cite{Su_2024_CVPR} clusters latent codes using $k$-means and conditions the generation process on cluster prototypes combined with label text prompts. MGD$^{3}$~\cite{santiagomgd} further discovers intra-class patterns through clustering to guide the denoising trajectory of the diffusion model. Inspired by this idea, VLCP~\cite{zou2025vlcp} enhances textual conditioning by leveraging LLaVA to generate per-image descriptions and selecting the most frequent sentence within each cluster to provide fine-grained textual descriptions for each cluster prototype. In these approaches, clustering is a crucial step for generating distilled data. However, clustering-based distillation often falls into a homogeneity trap. Because cluster prototypes frequently overlap in dense data regions, the resulting synthetic samples become repetitive, sacrificing dataset diversity for mere representative proximity.

\section{Method}
\begin{figure*}[t]
  \centering
   \resizebox{1.0\linewidth}{!}{
  \includegraphics[width=1.0\textwidth]{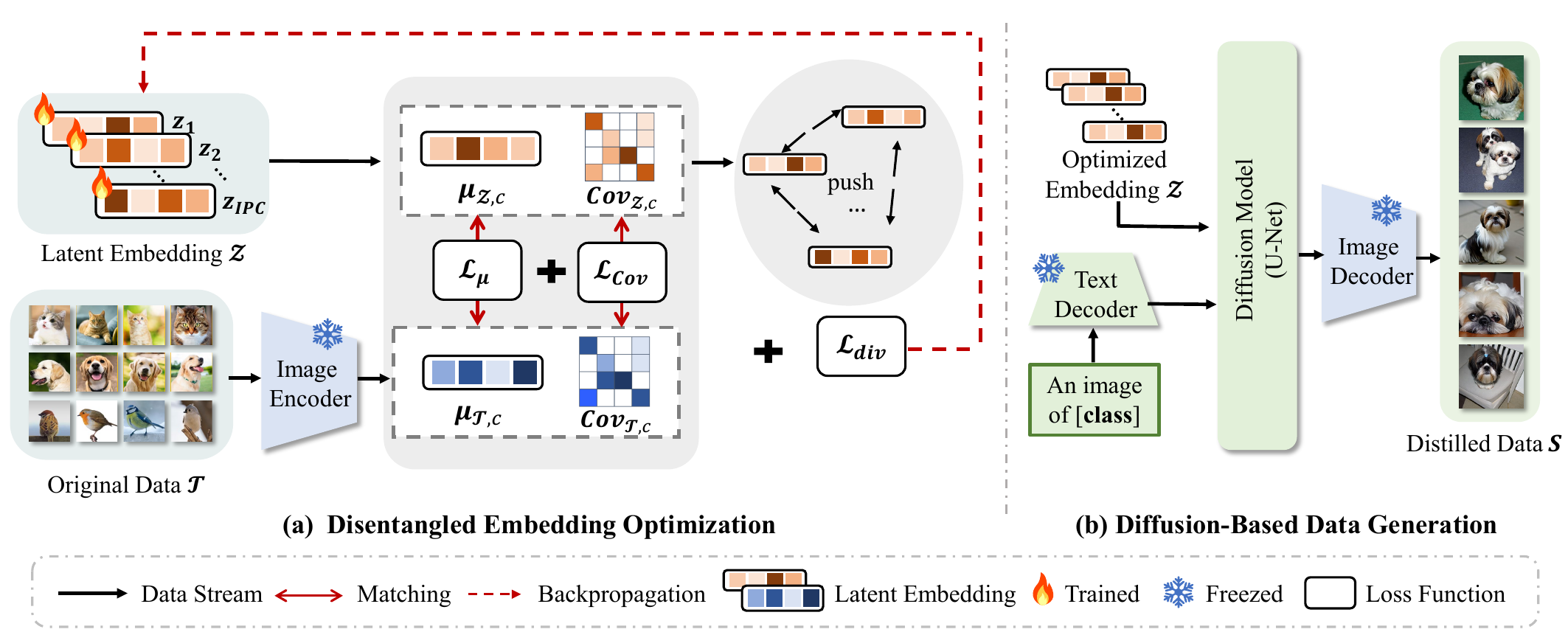}}
\caption{\textbf{Overview of our Disentangled Embedding Optimization (DEO) for diffusion-based distillation.} 
(\textbf{a}) Learnable latent embeddings $\mathcal{Z}_c$ are optimized to match the mean and covariance of the original data, with a diversity loss to encourage separation. 
(\textbf{b}) The optimized $\mathcal{Z}_c$ and text prompts are fed into the diffusion model to generate distilled samples $\mathcal{S}$.
}
  \label{dd_pipline}
\end{figure*}
\subsection{Preliminaries}
\label{sec:preliminaries}

Dataset distillation aims to construct a compact synthetic dataset
$\mathcal{S} = \{(\bm{s}_i, y_i)\}_{i=1}^{|\mathcal{S}|}$, 
on which models trained can achieve comparable generalization performance to those trained on the full dataset 
$\mathcal{T} = \{(\bm{x}_i, y_i)\}_{i=1}^{|\mathcal{T}|}$, 
where $y_i \in \mathcal{Y} = \{0, 1, \dots, C\!-\!1\}$. 
Typically, $\mathcal{S}$ contains only $\text{IPC} \times C$ samples 
($\text{IPC}$: images per class), satisfying $|\mathcal{S}| \ll |\mathcal{T}|$.
The goal is to train $\bm{\theta}_{\mathcal{S}}$ on $\mathcal{S}$ such that it achieves comparable generalization to $\bm{\theta}_{\mathcal{T}}$ trained on $\mathcal{T}$.

\textbf{Latent Diffusion Model for Dataset Distillation.}
Recently, latent diffusion models (LDM) have emerged as a promising paradigm for dataset distillation,
due to their strong cross-architecture generalization and efficient sample generation~\cite{Su_2024_CVPR, zou2025vlcp}.
A typical LDM pipeline consists of:
(1) a VAE image encoder $\mathcal{E}$ for mapping input images into the latent space,
(2) a conditioning text encoder that encodes class labels or descriptions,
(3) a denoising U-Net $\epsilon_\theta$ trained to reverse noise corruption,
and (4) a VAE decoder $\mathcal{D}$ that reconstructs images from the denoised latents.

\textbf{Latent Diffusion Objective.}
The latent diffusion model is trained to predict additive noise injected into latent embeddings:
\begin{equation}
\label{eq:ldm}
\mathcal{L}_{\mathrm{LDM}} =
\mathbb{E}_{\bm{x},\,y,\,\epsilon,\,t}
\left[\left\|\epsilon - \epsilon_\theta\left(\bm{z}_t, t, \tau_\phi(y)\right)\right\|_2^2\right],
\end{equation}
where $\bm{z}_t$ is a noised latent obtained from an original image $\bm{x}$ through a VAE encoder and a forward diffusion process, 
$\epsilon \sim \mathcal{N}(0, I)$ is the ground-truth noise, and $t \sim \mathcal{U}\{1, T\}$ is the diffusion timestep. 
The term $\tau_(y)$ denotes the embedding of the class label $y$ produced by a text encoder $\tau_\phi$, 
which provides semantic conditioning for the denoising network $\epsilon_\theta$. 
During inference, a scheduler progressively denoises the latent from $\bm{z}_T$ to $\bm{z}_0$,
which is then decoded into an image using the VAE decoder.
In the context of diffusion-based dataset distillation, 
the collection of such generated images constitutes the distilled dataset $\mathcal{S}$.

\subsection{Disentangled Embedding Optimization}

Recent diffusion-based dataset distillation methods typically perform clustering on the latent embeddings extracted by a VAE image encoder for each class and use the cluster prototypes as the inputs to the latent diffusion model to generate distilled images. However, since the samples of a class are often uniformly distributed in the latent space, the resulting cluster prototypes tend to lie close to one another. This will lead to the synthetic samples with limited intra-class diversity.

To mitigate this issue, instead of following the clustering-based pipeline, we propose to explicitly optimize the embeddings of distilled data $\mathcal{Z}_c = \{\bm{z}_i\}_{i=1}^{\textsc{IPC}}$. This design enables the generated samples to preserve the diversity of the original data distribution. As shown in Figure~\ref{dd_pipline}(a), we initialize the learnable latent embeddings $\mathcal{Z}_c$ independently and align them with the Gaussian distribution of the original data in the latent space of a pre-trained VAE encoder $\mathcal{E}$ by minimizing a mean alignment loss $\mathcal{L}_{\bm{\mu}}$ and a covariance alignment loss $\mathcal{L}_{\bm{Cov}}$. In addition, to promote sample-level diversity within each class, we introduce a diversity loss $\mathcal{L}_{div}$ that encourages the optimized embeddings to be mutually dispersed in the latent space.

Specifically, for each class $c$, let $\{\bm{e}_i = \mathcal{E}(\bm{x}_i)\}_{i=1}^{N_c}$ denote the latent embeddings extracted from the original samples in class $c$, where $N_c$ is the number of original samples. We define the mean alignment loss as follows:
\begin{equation}
\label{eq:mean_loss}
\begin{aligned}
\mathcal{L}_{\bm{\mu}} &= \left\| \bm{\mu}_{\mathcal{Z},c} - \bm{\mu}_{\mathcal{T},c} \right\|_2^2  \\
\emph{s.t.} \quad \bm{\mu}_{\mathcal{Z},c} &= \frac{1}{\textsc{IPC}} \sum_{i=1}^{\textsc{IPC}}\bm{z}_i, \quad \bm{\mu}_{\mathcal{T},c} = \frac{1}{N_c} \sum_{i=1}^{N_c} \bm{e}_i,
\end{aligned}
\end{equation}
where $\bm{\mu}_{\mathcal{Z},c}$ and $\bm{\mu}_{\mathcal{T},c}$ are the mean latent embeddings of the synthetic and original samples in class $c$, respectively.
Similarly, we define the covariance alignment loss as follows:
\begin{equation}
\label{eq:cov_loss}
\begin{aligned}
\mathcal{L}_{\bm{Cov}} &= \left\| \bm{Cov}_{\mathcal{Z},c} - \bm{Cov}_{\mathcal{T},c} \right\|_2^2\\
  \emph{s.t.}  \quad \bm{Cov}_{\mathcal{Z},c} &= \frac{1}{\textsc{IPC}} \sum_{i=1}^{\textsc{IPC}} (\bm{z}_i - \bm{\mu}_{\mathcal{Z},c})(\bm{z}_i - \bm{\mu}_{\mathcal{Z},c})^\top, \\
\bm{Cov}_{\mathcal{T},c} &= \frac{1}{N_c} \sum_{i=1}^{N_c} (\bm{e}_i - \bm{\mu}_{\mathcal{T},c})(\bm{e}_i - \bm{\mu}_{\mathcal{T},c})^\top.
\end{aligned}
\end{equation}

To further encourage sample-level diversity within each class, we introduce a separation regularization term that penalizes latent embeddings with high similarity:
\begin{equation}
\label{eq:div}
\mathcal{L}_{div} = \frac{1}{\textsc{IPC}} \sum_{i=1}^{\textsc{IPC}} \sum_{\substack{j=1 \\ j \neq i}}^{\textsc{IPC}} \log\left(1 + \exp\left(\tilde{\bm{z}}_i^\top \tilde{\bm{z}}_j\right)\right),
\end{equation}
where $\tilde{\bm{z}}_i$ denotes the $\ell_2$-normalized version of $\bm{z}_i$. This term encourages the learnable embeddings to be well separated in the latent space, thereby promoting intra-class diversity in the distilled data.

\begin{algorithm}[t]
\small
\caption{\textbf{DEO}}
\label{alg:distillation}
\KwIn{
$\mathcal{T}$: Original dataset; 
$C$: Number of classes; 
$\mathrm{IPC}$: Images per class; 
$T$: Number of optimization steps; 
$\eta$: Learning rate.
}

\BlankLine
\underline{\textbf{(a): Disentangled Embedding Optimization}} \\
\For{each class $c = 1$ \KwTo $C$}{
    Initialize learnable latent embeddings $\mathcal{Z}_c = \{\bm{z}_i\}_{i=1}^{\mathrm{IPC}}$. \\
    Compute the class-wise feature mean $\bm{\mu}_{\mathcal{T},c}$ and covariance $\bm{Cov}_{\mathcal{T},c}$ from  $\mathcal{T}$. \\
    \For{iteration $t = 1$ \KwTo $T$}{
        Compute the mean $\bm{\mu}_{\mathcal{Z},c}$ and covariance $\bm{Cov}_{\mathcal{Z},c}$ of the latent embeddings $\mathcal{Z}_c$. \\
        Compute the mean  loss $\mathcal{L}_{\bm{\mu}}$ via Eq.~\eqref{eq:mean_loss}. \\ 
        Compute the covariance  loss $\mathcal{L}_{\bm{Cov}}$ via Eq.~\eqref{eq:cov_loss}. \\ 
        Compute the diversity regularization loss $\mathcal{L}_{\text{div}}$ via Eq.~\eqref{eq:div}. \\
        Compute the overall objective $\mathcal{L}$ via Eq.~\eqref{eq:total_loss}. \\
        Update latent embeddings: 
        $\mathcal{Z}_c \leftarrow \mathcal{Z}_c - \eta \nabla_{\mathcal{Z}_c}\mathcal{L}$. \\
    }
}

\BlankLine
\underline{\textbf{(b): Diffusion-Based Data Generation}} \\
\For{each class $c = 1$ \KwTo $C$}{
    Combine the optimized latent embeddings $\mathcal{Z}_c$ with the class-specific textual prompt $P_{\text{text}}^{c}$ (i.e., ``\texttt{An image of [class]}''). \\
    Generate synthetic images $\{\bm{s}_i\}$ according to Eq.~\eqref{eq:generation}. \\
    Append generated images to the distilled dataset: 
    $\mathcal{S} \leftarrow \mathcal{S} \cup \{\bm{s}_i\}$. \\
}

\KwOut{Distilled synthetic dataset $\mathcal{S}$.}
\end{algorithm}

\subsection{Overall Objective and Generation Procedure}
The training objective combines distribution alignment with diversity regularization:
\begin{equation}
\label{eq:total_loss}
\mathcal{L} \;= \mathcal{L}_{\bm{\mu}} + \lambda_{1}\,\mathcal{L}_{\bm{Cov}} \;+ \lambda_{2}\,\mathcal{L}_{div},
\end{equation}
where $\lambda_{1}$ and $\lambda_{2}$ are tunable hyperparameters.

After the latent embeddings are optimized, we feed them into the diffusion model following prior work~\cite{Su_2024_CVPR,zou2025vlcp}. As illustrated in Figure~\ref{dd_pipline}(b), the per-class distilled latent set $\mathcal{Z}_c=\{\bm{z}_i\}_{i=1}^{\textsc{IPC}}$ is paired with a class-specific textual prompt $P_{\text{text}}^{c}$ (e.g., ``\texttt{An image of [class]}''). The prompt embedding $\tau_{\phi}(P_{\text{text}}^{c})$ conditions the U-Net, and each distilled image $\bm{s}_i$ is synthesized as:
\begin{equation}
\label{eq:generation}
\bm{s}_i
=
\mathcal{D}\!\left(
  \mathcal{U}_{t}\!\left(
    \operatorname{Concat}\!\left(\tilde{\bm{z}}^{\,c}_{i,t},\;
    \tau_{\phi}\!\left(P_{\text{text}}^{c}\right)\right)
  \right)
\right),
\end{equation}
where $\tilde{\bm{z}}^{\,c}_{i,t}$ is the noise-perturbed version of $\bm{z}_i$ at diffusion timestep $t$, 
$\mathcal{U}_{t}(\cdot)$ denotes the timestep-dependent U\textminus Net backbone, 
$\mathcal{D}(\cdot)$ is the image decoder, and $\operatorname{Concat}(\cdot,\cdot)$ denotes channel-wise concatenation. This formulation enables the optimized latent codes and textual semantics to jointly guide the denoising process to generate the distilled dataset $\mathcal{S}$.  The complete training and generation pipeline is summarized in Algorithm~\ref{alg:distillation}.

\definecolor{lightgray}{RGB}{219,219,219}
\newcommand{\xpm}[1]{_{\text{\small$\pm$\,#1}}}

\begin{table*}[htbp]
\small
\centering
\caption{Performance comparison on ImageWoof, ImageNette, and ImageIDC with ResNetAP-10 under different IPC settings. The best and second-best results are highlighted in \textbf{bold} and \underline{underline}, respectively; ``--'' denotes unavailable results.}
\label{tab:rn10_three_datasets}
\setlength\tabcolsep{8.2pt}
 \vskip-2pt
\begin{tabular}{l|ccc|ccc|ccc}
\toprule
\multirow{2}{*}{Method} 
& \multicolumn{3}{c|}{ImageWoof} 
& \multicolumn{3}{c|}{ImageNette} 
& \multicolumn{3}{c}{ImageIDC} \\
\cmidrule(lr){2-4} \cmidrule(lr){5-7} \cmidrule(lr){8-10}
& IPC=10 & IPC=20 & IPC=50 
& IPC=10 & IPC=20 & IPC=50 
& IPC=10 & IPC=20 & IPC=50 \\
\midrule
Random   & 29.4$\pm$0.8 & 32.7$\pm$0.4 & 47.2$\pm$1.3 & 54.2$\pm$1.6 & 63.5$\pm$0.5 & 76.1$\pm$1.1 & 48.1$\pm$0.8 & 52.5$\pm$0.9 & 68.1$\pm$0.7 \\
K-Center & 22.1$\pm$0.1 & 25.1$\pm$0.7 & 40.6$\pm$0.4 & -- & -- & -- & -- & -- & -- \\
Herding  & 32.0$\pm$0.3 & 34.9$\pm$0.1 & 49.1$\pm$0.7 & -- & -- & -- & -- & -- & -- \\
DiT      & 34.7$\pm$0.5 & 41.1$\pm$0.8 & 49.3$\pm$0.2 & 59.1$\pm$0.7 & 64.8$\pm$1.2 & 73.3$\pm$0.9 & 54.1$\pm$0.4 & 58.9$\pm$0.2 & 64.3$\pm$0.6 \\
DM       & 30.3$\pm$1.2 & 35.2$\pm$0.6 & 47.1$\pm$1.1 & 60.8$\pm$0.6 & 66.5$\pm$1.1 & 76.2$\pm$0.4 & 52.8$\pm$0.5 & 58.5$\pm$0.4 & 69.1$\pm$0.8 \\
IDC-1    & 39.1$\pm$0.5 & 43.4$\pm$0.3 & 48.3$\pm$1.0 & -- & -- & -- & -- & -- & -- \\
Minimax  & 36.2$\pm$3.2 & 43.3$\pm$2.7 & 53.9$\pm$0.7 & 57.7$\pm$1.2 & 64.7$\pm$0.8 & 73.9$\pm$0.3 & 51.9$\pm$1.4 & 59.1$\pm$3.7 & 69.4$\pm$1.4 \\
D$^4$M   & 33.2$\pm$2.1 & 40.1$\pm$1.6 & 51.7$\pm$3.2 & 60.9$\pm$1.7 & 66.3$\pm$1.3 & 77.7$\pm$1.1 & 50.3$\pm$1.0 & 55.8$\pm$0.2 & 69.1$\pm$2.4 \\
VLCP     & \underline{39.5$\pm$1.5} & \underline{44.5$\pm$2.2} & \underline{57.3$\pm$0.5} & \underline{64.8$\pm$3.6} & \underline{71.4$\pm$0.5} & \underline{81.2$\pm$0.8} & \underline{57.0$\pm$1.4} & \underline{63.3$\pm$1.2} & \underline{71.9$\pm$0.4} \\
\rowcolor{lightgray}
\textbf{DEO (Ours)} 
& \textbf{40.9$\pm$0.5} & \textbf{46.7$\pm$0.4} & \textbf{59.9$\pm$0.6}
& \textbf{66.2$\pm$1.2} & \textbf{72.6$\pm$0.4} & \textbf{82.2$\pm$0.4}
& \textbf{58.6$\pm$0.7} & \textbf{64.6$\pm$0.6} & \textbf{72.6$\pm$0.3} \\
\bottomrule
\end{tabular}
 \vskip-2pt
\end{table*}

\section{Discussion}

The advantage of DEO over clustering-based latent distillation methods can be understood from the perspective of prototype bias caused by boundary ambiguity. In the latent space of a pretrained VAE, samples within the same class are often uniformly distributed without clear sub-cluster boundaries. As a result, samples near the overlapping region may be arbitrarily assigned to different clusters, which will pull the estimated cluster prototypes near each other.

Consider two ideal semantic modes with centers $\mu_1$ and $\mu_2$. Due to boundary ambiguity, the estimated prototypes can be written as
\begin{equation}
\hat{\mu}_1 = (1-\alpha)\mu_1 + \alpha \mu_2,\qquad
\hat{\mu}_2 = (1-\beta)\mu_2 + \beta \mu_1,
\end{equation}
where $\alpha,\beta \in [0,0.5)$ quantify cross-mode contamination. Their distance then becomes
\begin{equation}
\|\hat{\mu}_1-\hat{\mu}_2\|_2
= (1-\alpha-\beta)\|\mu_1-\mu_2\|_2
< \|\mu_1-\mu_2\|_2.
\end{equation}
This simple analysis shows that boundary ambiguity naturally shrinks the separation between cluster prototypes, causing the diffusion model to receive  similar latent inputs. In contrast, DEO avoids hard cluster assignment and directly optimizes learnable embeddings under distribution matching and dispersion regularization, thereby preserving richer intra-class semantic variability.

\section{Experiment}

\begin{table}[t]
\centering
\caption{Performance comparison on ImageNet-1K with IPC=10/50.
}
\setlength\tabcolsep{2.0pt}
\small
\begin{tabular}{l|ccccc>{\columncolor{lightgray}}c}
\toprule
IPC & SRe2L & RDED & DiT & Minimax & VLCP & \textbf{DEO (Ours)} \\
\midrule
10 & 21.3$\pm$0.6 & 42.0$\pm$0.1 & 39.6$\pm$0.4 & 44.3$\pm$0.5 & 46.7$\pm$0.4 & \textbf{49.2$\pm$0.8} \\
50 & 46.8$\pm$0.2 & 56.5$\pm$0.1 & 52.9$\pm$0.6 & 58.6$\pm$0.3 & 60.5$\pm$0.2 & \textbf{61.4$\pm$0.3} \\
\bottomrule
\end{tabular}
\label{tab:imagenet1k_only_transposed}
   \vspace{-0.7cm}
\end{table}

\subsection{Experimental Setup}

\textbf{Datasets.} We evaluate our proposed DEO method on a diverse suite of image benchmarks. For low-resolution settings, we utilize CIFAR-10 and CIFAR-100 \cite{krizhevsky2009learning}, two widely used datasets containing 60,000 images at $32 \times 32$ resolution, with 10 and 100 classes respectively. For high-resolution evaluation, we employ ImageNet-1K \cite{deng2009imagenet}, a large-scale dataset comprising 1.28 million images across 1,000 categories at $224 \times 224$ resolution, which offers a comprehensive test of generalization under diverse visual concepts. Additionally, we assess performance on several widely adopted subsets of ImageNet-1K, specifically ImageWoof, ImageNette \cite{imagenette_howard_2019}, and ImageIDC \cite{zou2025vlcp}. ImageWoof focuses on fine-grained recognition among ten visually similar dog breeds, while ImageNette and ImageIDC serve as more general benchmarks with ten coarse-grained categories each.

\textbf{Implementation Details.} We follow the experimental protocols established in prior work~\cite{Su_2024_CVPR, zou2025vlcp}. Unless otherwise specified, we fine-tune Stable Diffusion v1.5 on each dataset using paired image-label supervision. The trainable latent embeddings are optimized using the Adam optimizer with a learning rate of $1 \times 10^{-3}$ for 10,000 steps. To ensure both distribution alignment and sample diversity, we incorporate a covariance matching loss weighted by $\lambda_{1} = 0.5$, alongside a diversity-promoting separation regularizer applied to the denoised latents with a coefficient of $\lambda_{2} = 0.004$. All experiments are conducted on NVIDIA RTX A6000 or NVIDIA A100 (80GB) GPUs. 


\textbf{Evaluation Protocol.} For a fair comparison, we follow the evaluation settings used in VLCP~\cite{zou2025vlcp}. Two evaluation strategies are employed: (1) Direct Training: a classifier is trained from scratch using the generated images and their labels, with performance evaluated on the standard test split. This setting is used for ImageWoof, ImageNette, and ImageIDC. (2) Soft-Label Supervision: soft labels are obtained by applying the teacher model pre-trained by the original data to the distilled images, which are then used to train a new classifier. This setting is adopted for ImageNet-1K and CIFAR-10/100. All reported Top-1 accuracies are averaged over three random seeds.

\textbf{Compared Methods.} We compare DEO against a comprehensive set of state-of-the-art baselines: (1) Core-set selection methods, including K-Center~\cite{sener2018active} and Herding~\cite{10.1145/1553374.1553517}; (2) Matching-based distillation methods, such as DM~\cite{zhao2023distribution}, IDC-1~\cite{kim2022dataset} SRe2L~\cite{yin2023squeeze}, and RDED~\cite{sun2024diversity}; and (3) Diffusion-based distillation methods, including DiT~\cite{peebles2023scalable}, Minimax~\cite{gu2024efficient}, D$^4$M~\cite{Su_2024_CVPR}, and VLCP~\cite{zou2025vlcp}.
\subsection{Main Results}
\textbf{ImageWoof, ImageNette, and ImageIDC.}
Table~\ref{tab:rn10_three_datasets} reports the results on three benchmarks under a unified ResNetAP-10 evaluation protocol.
DEO consistently achieves the best performance across all datasets and all IPC settings.
On ImageWoof, DEO reaches 40.9\%, 46.7\%, and 59.9\% at IPC=10, 20, and 50, respectively, outperforming the strongest baseline VLCP by 1.4\%, 2.2\%, and 2.6\%.
Similarly, on ImageNette, DEO improves over VLCP by 1.4\%, 1.2\%, and 1.0\%, while on ImageIDC, the gains are 1.6\%, 1.3\%, and 0.7\%.
These results show that DEO produces more effective distilled samples across datasets with different visual characteristics and compression ratios.

\textbf{ImageNet-1K.}
Table~\ref{tab:imagenet1k_only_transposed} further demonstrates the scalability of DEO on the challenging ImageNet-1K benchmark.
DEO achieves 49.2\% at IPC=10 and 61.4\% at IPC=50, surpassing the strongest baseline VLCP by 2.5\% and 0.9\%, respectively.
The consistent advantage on ImageNet-1K indicates that our embedding optimization strategy remains effective even when the dataset scale and semantic diversity increase substantially.

\textbf{CIFAR-10/100.}
We further evaluate DEO on low-resolution benchmarks in Table~\ref{tab:cifar}.
On CIFAR-10, DEO achieves 41.8\% and 64.1\% at IPC=10 and 50, improving over VLCP by 2.8\% and 0.9\%, respectively.
On CIFAR-100, DEO reaches 52.4\% and 66.8\%, exceeding VLCP by 1.8\% and 0.7\%.
Compared with earlier baselines, the gains are even more pronounced; for example, on CIFAR-100 with IPC=10, DEO outperforms SRe2L and RDED by 25.4\% and 9.8\%, respectively.
Overall, these results demonstrate that DEO consistently delivers superior distilled data quality on both large-scale high-resolution benchmarks and standard low-resolution datasets.

\subsection{Cross-architecture Evaluation} Table~\ref{tab:imagewoof_conv_rn18_transposed} reports the results on ImageWoof using two target architectures, ConvNet and ResNet-18.
DEO consistently achieves the best performance across all IPC settings on both backbones, demonstrating strong cross-architecture generalization of the distilled data.
Specifically, on ConvNet, DEO improves over the strongest baseline VLCP by 1.4\%, 1.5\%, and 1.1\% at IPC=10, 20, and 50, respectively.
On ResNet-18, the corresponding gains are 1.6\%, 0.5\%, and 1.2\%.
Moreover, compared with the most relevant diffusion-based baseline D$^4$M, DEO achieves substantially larger gains, ranging from 5.4\% to 9.2\% across different settings.
These results indicate that DEO learns more transferable and discriminative latent representations, leading to superior performance across different model families rather than on a single evaluation architecture only.

\subsection{Analysis} 
\textbf{Ablation Study. } 
To evaluate the effectiveness of each component in our DEO, we conduct ablation studies on ImageWoof under three IPC settings: IPC = 10, 20, and 50. We analyze the impact of three core losses: mean alignment ($\mathcal{L}_{\bm{\mu}}$), covariance alignment ($\mathcal{L}_{\bm{Cov}}$), and diversity regularization ($\mathcal{L}_{div}$). As summarized in Table~\ref{tab:ablation_study}, we can draw four key conclusions. First, using only $\mathcal{L}_{\bm{\mu}}$ already outperforms D$^4$M by 1.1\% and 0.5\%, demonstrating that aligning the class-wise means of latent embeddings effectively preserves the core semantics of the original data. Second, adding $\mathcal{L}_{\bm{Cov}}$ further captures intra-class variations around the class centers and brings the performance up to par with VLCP. This indicates that matching both the mean and covariance statistics is sufficient to reconstruct rich semantic information without relying on large pretrained models or long-horizon reasoning. Third, incorporating $\mathcal{L}_{div}$ alongside $\mathcal{L}_{\bm{\mu}}$ yields an additional gain of 1.9\%–2.5\%, suggesting that enforcing separation among embeddings promotes greater semantic diversity and expressiveness. Finally, the combination of all three losses consistently achieves the best performance across all IPC settings, validating the effectiveness of jointly optimizing for semantic alignment and intra-class diversity in the latent embedding space.

\begin{table}[t]
\setlength\tabcolsep{4.0pt}
\small
\centering
\caption{Performance comparison on CIFAR-10 and CIFAR-100 with IPC=10/50.}
\begin{tabular}{lcccc>{\columncolor{lightgray}}c}
\toprule
Dataset      & IPC & SRe2L       & RDED           & VLCP     & \textbf{DEO (Ours)}              \\ \midrule
\multirow{2}{*}{CIFAR-10} & 10  & 29.3$\pm$0.5 & 37.1$\pm$0.3 & 39.0$\pm$0.7 & \textbf{41.8$\pm$0.6} \\

         & 50  & 45.0$\pm$0.7 & 62.1$\pm$0.1 & 63.2$\pm$0.3 &\textbf{64.1$\pm$0.7}  \\ \midrule
\multirow{2}{*}{CIFAR-100} & 10  & 27.0$\pm$0.4 & 42.6$\pm$0.2 & 50.6$\pm$0.7 &\textbf{52.4$\pm$0.3} \\
                           & 50  & 50.2$\pm$0.4 & 62.6$\pm$0.1 & 66.1$\pm$0.3 &\textbf{66.8$\pm$0.5} \\ \bottomrule
\end{tabular}
\label{tab:cifar}
   \vspace{-0.4cm}
\end{table}

\begin{table*}[t]
\small
\centering
\caption{Performance comparison of dataset distillation methods on ImageWoof under varying IPC values.
Evaluations are conducted using ConvNet and ResNet-18.
The best results are shown in \textbf{bold}.}
\label{tab:imagewoof_conv_rn18_transposed}
\setlength\tabcolsep{4.9pt}
\begin{tabular}{cc|ccccccccc>{\columncolor{lightgray}}c}
\toprule
Backbone & IPC & Random & K-Center & Herding & DiT & DM & IDC-1 & Minimax & D$^4$M & VLCP & \textbf{DEO (Ours)} \\
\midrule
\multirow{3}{*}{ConvNet}
& 10 & 24.3$\pm$1.1 & 19.4$\pm$0.9 & 26.7$\pm$0.5 & 34.2$\pm$1.1 & 26.9$\pm$1.2 & 33.3$\pm$1.1 & 33.3$\pm$1.7 & 29.4$\pm$0.9 & 34.8$\pm$2.4 & \textbf{36.2$\pm$0.8} \\
& 20 & 29.1$\pm$0.7 & 21.5$\pm$0.8 & 29.5$\pm$0.3 & 36.1$\pm$0.8 & 29.9$\pm$1.0 & 35.5$\pm$0.8 & 37.3$\pm$0.1 & 34.0$\pm$2.3 & 37.9$\pm$1.9 & \textbf{39.4$\pm$0.3} \\
& 50 & 41.3$\pm$0.6 & 36.5$\pm$1.0 & 40.3$\pm$0.7 & 46.5$\pm$0.8 & 44.4$\pm$1.0 & 43.9$\pm$1.2 & 50.9$\pm$0.8 & 47.4$\pm$0.9 & 54.5$\pm$0.6 & \textbf{55.6$\pm$0.3} \\
\midrule
\multirow{3}{*}{ResNet-18}
& 10 & 27.7$\pm$0.9 & 21.1$\pm$0.4 & 30.2$\pm$1.2 & 34.7$\pm$0.4 & 33.4$\pm$0.7 & 37.3$\pm$0.2 & 35.7$\pm$1.6 & 32.3$\pm$1.2 & 39.9$\pm$2.6 & \textbf{41.5$\pm$1.1} \\
& 20 & 29.7$\pm$0.5 & 23.6$\pm$0.3 & 32.2$\pm$0.6 & 40.5$\pm$0.5 & 29.8$\pm$1.7 & 38.6$\pm$0.2 & 41.8$\pm$1.9 & 38.4$\pm$1.1 & 44.5$\pm$2.0 & \textbf{45.0$\pm$0.8} \\
& 50 & 47.9$\pm$1.8 & 39.6$\pm$1.0 & 48.3$\pm$1.2 & 50.1$\pm$0.5 & 46.2$\pm$0.6 & 48.3$\pm$0.8 & 53.7$\pm$0.6 & 53.7$\pm$2.2 & 58.9$\pm$1.5 & \textbf{60.1$\pm$0.9} \\
\bottomrule
\end{tabular}
\end{table*}

\begin{table}[t]
\centering
\setlength{\tabcolsep}{3.6pt} 
\caption{Ablation study on ImageWoof.}
\small
\captionsetup{font=small}
\begin{tabular}{c|ccc|ccc}
\toprule
 \multirow{1.25}{*}{Method} 
 &$\mathcal{L}_{\bm{\mu}}$ &  $\mathcal{L}_{\bm{Cov}}$  &$\mathcal{L}_{div}$ &IPC=10 &IPC=20  &IPC=50  \\\midrule 
D$^{4}$M  &- &-  &- &33.2  &40.1 &51.7  \\\midrule
\multirow{4}{*}{Ours} &{\Checkmark} &-  &- &34.3  &39.6 &52.2 \\
&{\Checkmark} & {\Checkmark}  &-       &39.8  &44.3 &57.8 \\
&{\Checkmark} & - &{\Checkmark}   &36.2   &42.1  &54.1 \\
&{\Checkmark} & {\Checkmark}  &{\Checkmark}  &40.9  &46.7 &59.9\\

\bottomrule
\end{tabular}
\label{tab:ablation_study}
\end{table}

\begin{figure} [t]
  \centering
    \includegraphics[width=0.8\linewidth]{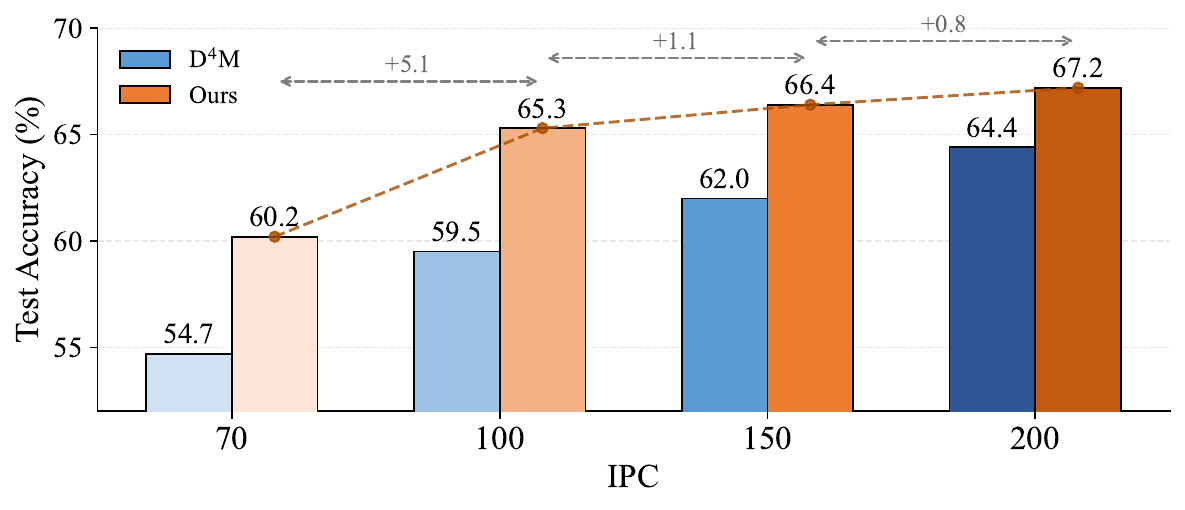}
     \vskip-2pt
\caption{Performance of D$^4$M and DEO under different high IPC settings on ImageWoof.}
   \vspace{-0.2cm}
 \label{ipc_more}
\end{figure}

\textbf{Effect of High IPC. } We further analyze the effect of high IPC settings in Figure~\ref{ipc_more} by comparing D$^4$M and DEO on ImageWoof. As IPC increases, both methods achieve better accuracy, showing that a larger distilled budget generally improves downstream performance. DEO consistently surpasses D$^4$M across all IPC settings, indicating its stronger ability to preserve informative class semantics. At the same time, the marginal gain of DEO gradually decreases as IPC grows, suggesting a clear diminishing-return effect. This implies that once sufficient semantic coverage is achieved, further increasing IPC brings only limited additional improvement.

\begin{figure} [t]
  \centering
    \includegraphics[width=0.95\linewidth]{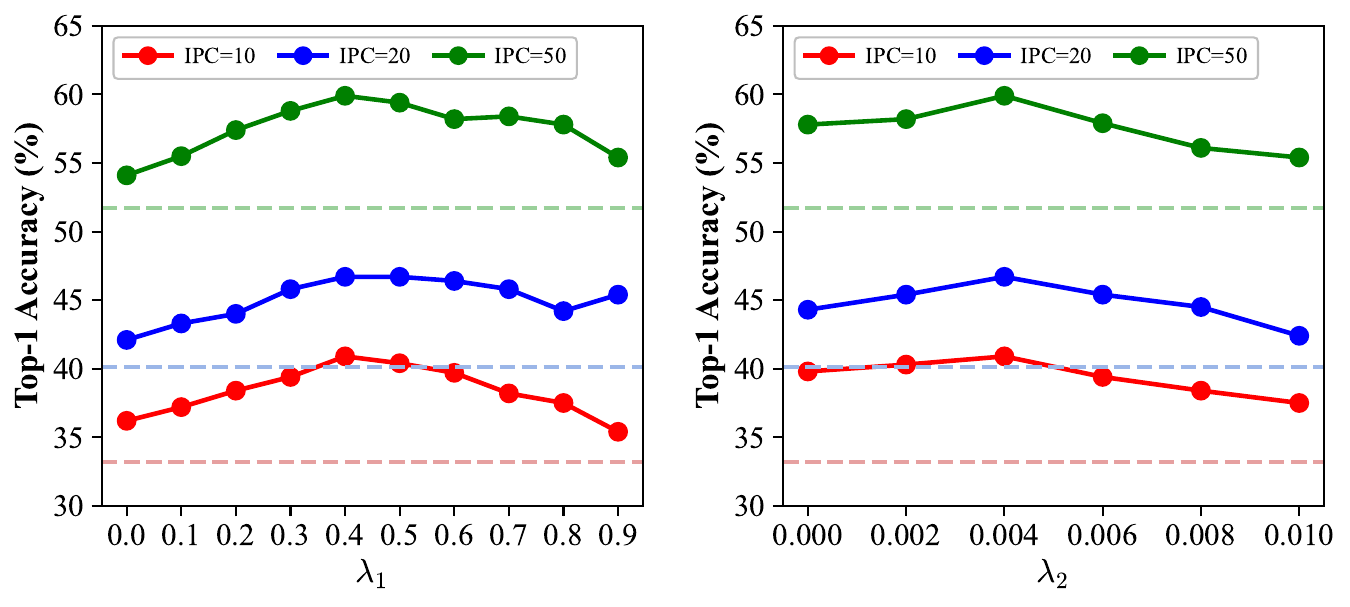}
\caption{Effect of $\lambda_1$ and $\lambda_2$ on ImageWoof with $\text{IPC} \in \{10, 20, 50\}$. Red, blue, and green lines represent $\text{IPC} = 10, 20,$ and $50$, respectively, while dashed lines denote the corresponding results from D$^4$M.}
   \vspace{-0.3cm}
 \label{fig_0.1}
\end{figure}

\textbf{Effect of $\lambda_1$. } 
In our DEO framework, the hyperparameter $\lambda_1$ controls the weight of the covariance alignment loss $\mathcal{L}_{\bm{Cov}}$, which is designed to match the second-order statistics of the learnable embeddings to those of the original data. Covariance reflects both intra-class dispersion and inter-dimensional correlations, thereby capturing semantic variations around the class center. As shown in the left panel of Figure~\ref{fig_0.1}, on ImageWoof with $\text{IPC} \in \{10, 20, 50\}$, increasing $\lambda_1$ from 0 to 0.5 consistently improves classification performance, demonstrating the effectiveness of incorporating higher-order statistical alignment. However, when $\lambda_1$ exceeds 0.5, the performance drops noticeably. We attribute this decline to the trade-off between mean alignment and covariance alignment: an overly large $\lambda_1$ may overemphasize noisy covariance estimates, especially in low-IPC settings, and drive the embeddings to fit outlier samples. Unless otherwise specified, we set $\lambda_1 = 0.5$ in all experiments.

\textbf{Effect of $\lambda_2$. } 
The hyperparameter $\lambda_2$ controls the strength of the diversity regularization, which encourages the learned embeddings to be more dispersed in the latent space. Such dispersion helps the diffusion model generate synthetic samples that are both more diverse and more representative. As shown in the right panel of Figure~\ref{fig_0.1}, we conduct a sensitivity analysis on ImageWoof with $\text{IPC} \in \{10, 20, 50\}$. The results indicate that performance reaches its peak at $\lambda_2 = 0.004$. When $\lambda_2$ becomes excessively large, however, the embeddings become over-dispersed, weakening intra-class compactness and leading to a gradual performance drop.

\begin{figure}[t]
  \centering
    \includegraphics[width=0.95\linewidth]{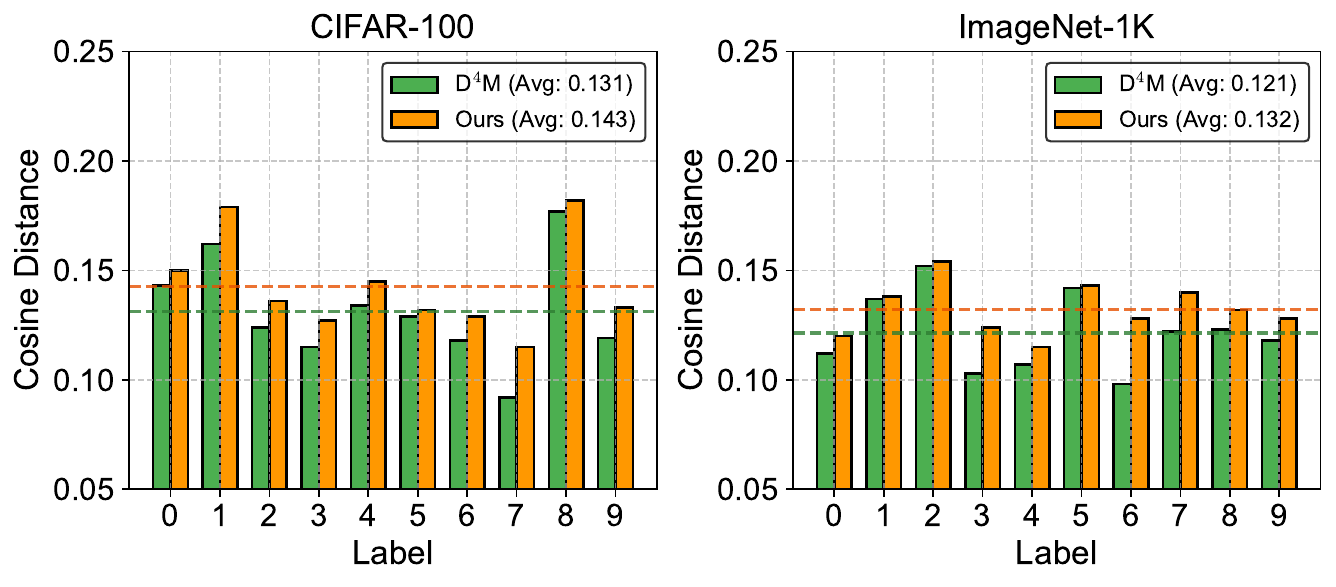}
  \caption{Comparison of intra-class diversity. Average cosine distance between distilled samples and class prototypes on CIFAR-100 and ImageNet-1K.}
   \vspace{-0.4cm}
  \label{distance_imagewoof}
\end{figure}

\begin{table}[t]
\setlength{\tabcolsep}{6.6pt} 
\centering
\caption{Per-class computational cost (minutes) at IPC=10 on CIFAR-100 and ImageNet-1K.}
\label{tab:comp_cost_vertical}
\small
\begin{tabular}{@{}lcccc@{}}
\toprule
\multirow{2}{*}{Stage (min)} & \multicolumn{2}{c}{CIFAR-100} & \multicolumn{2}{c}{ImageNet-1K} \\
\cmidrule(lr){2-3} \cmidrule(lr){4-5}
& VLCP & DEO (Ours) & VLCP & DEO (Ours) \\
\midrule
Clustering      & 1.3  & --            & 3.5  & --            \\
LLaVA Inf.      & 8.2  & --            & 21.7 & --            \\
Emb. Stats      & --   & 3.6           & --   & 9.6           \\
Optim.          & --   & 3.8           & --   & 3.8           \\
\midrule
Total Time & 9.5  & 7.4  & 25.2 & 13.4 \\
\bottomrule
\end{tabular}
\end{table}
\begin{figure}[t]
  \captionsetup[subfigure]{font=tiny} 
  \centering
  \resizebox{1.0\linewidth}{!}{%
  \begin{subfigure}[b]{0.245\linewidth}
    \centering
    \includegraphics[width=1.0\linewidth]{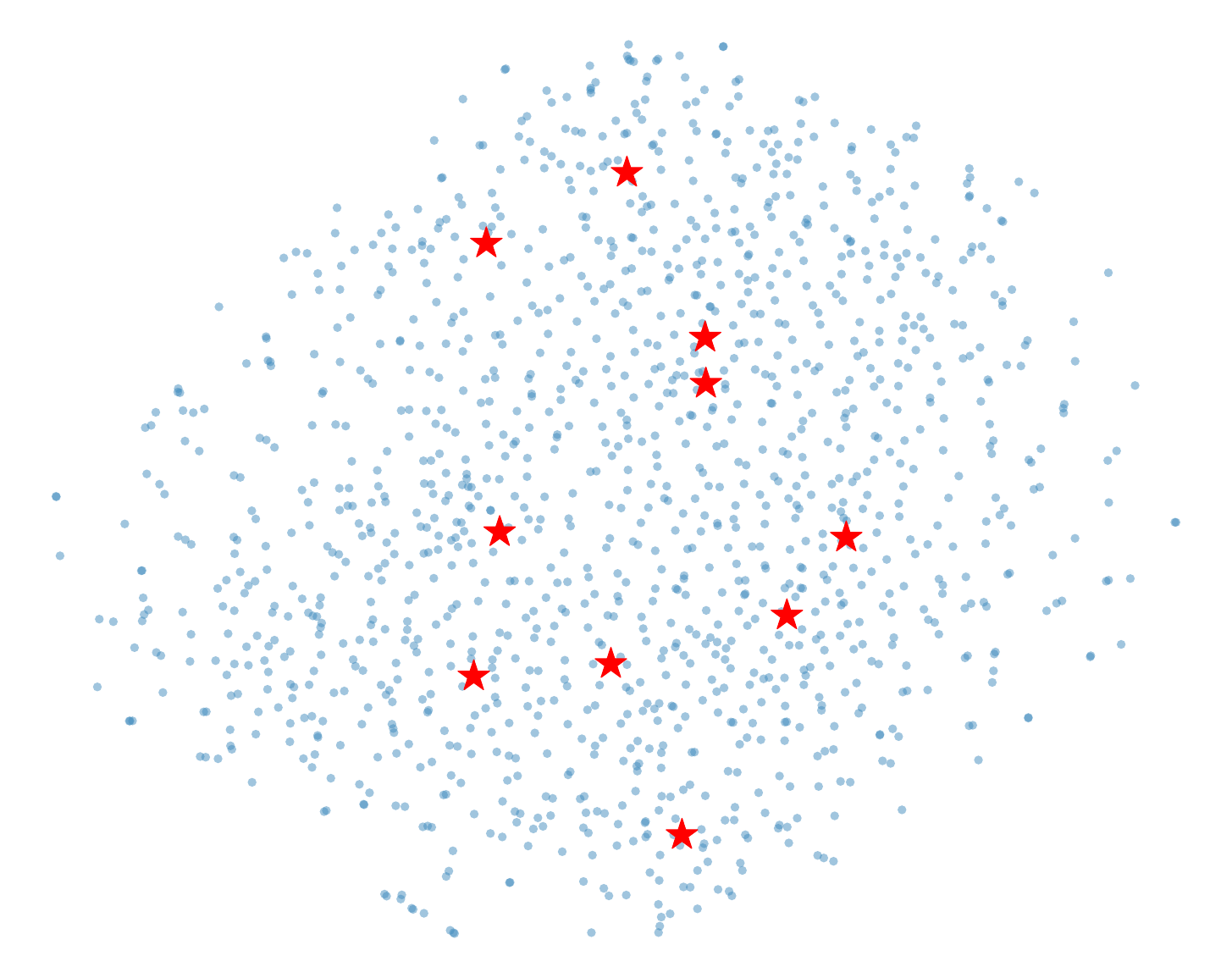} 
    \subcaption{D$^4$M (IPC=10)}
    \label{mcp:1}
  \end{subfigure}
  \begin{subfigure}[b]{0.245\linewidth}
    \centering
    \includegraphics[width=1.0\linewidth]{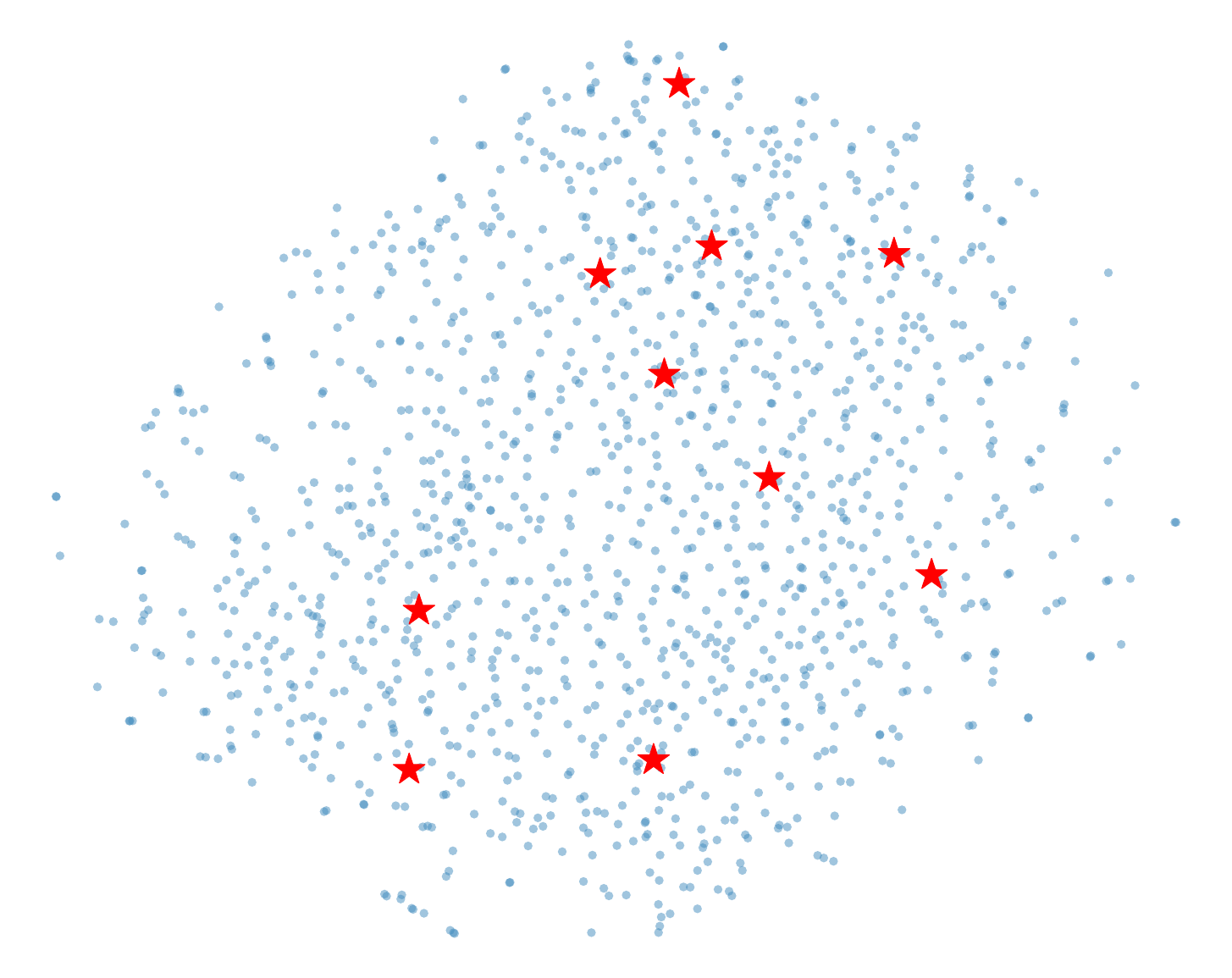} 
      \subcaption{DEO (IPC=10)}
    \label{mcp:2}
  \end{subfigure} 
  \begin{subfigure}[b]{0.245\linewidth}
    \centering
    \includegraphics[width=1.0\linewidth]{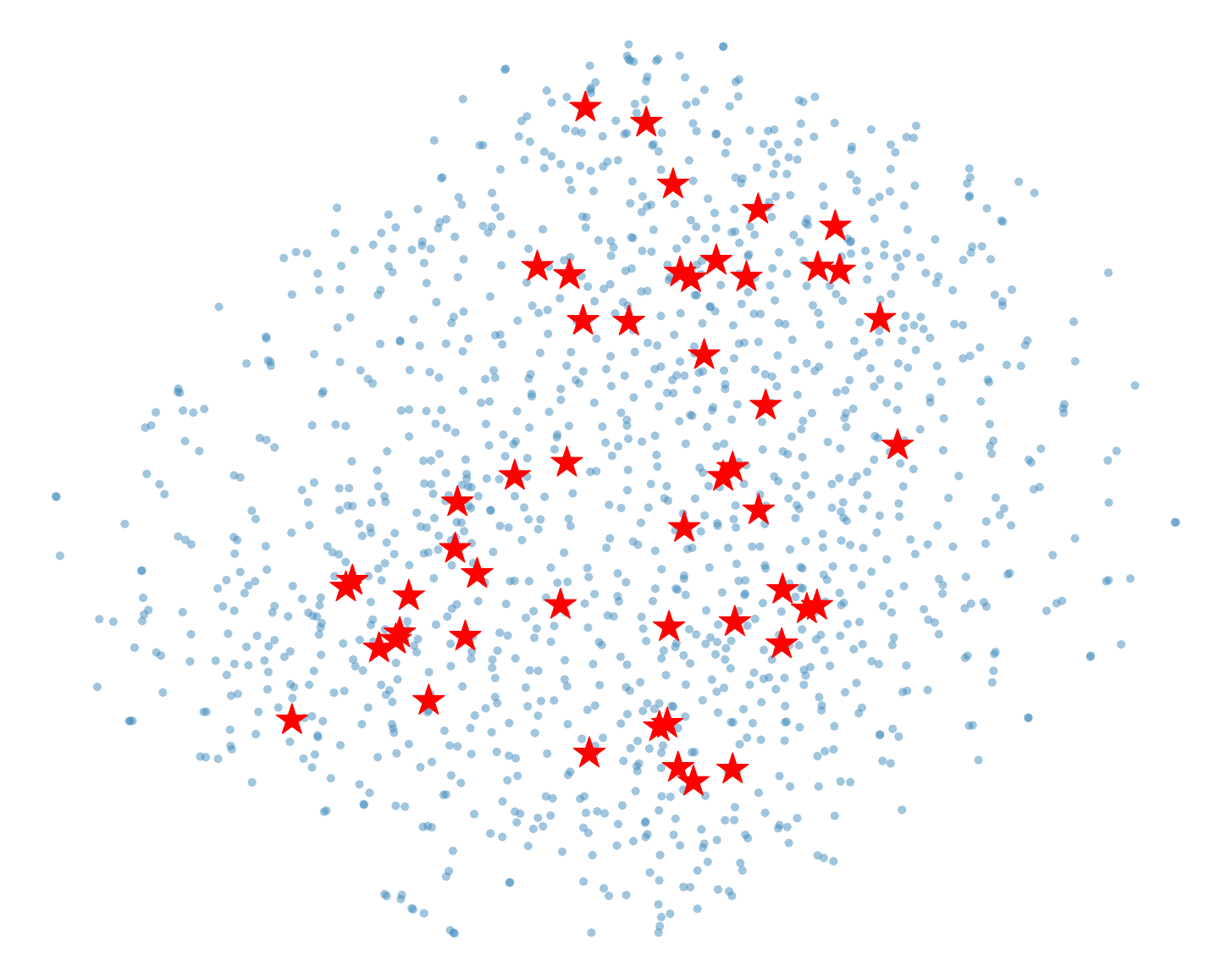} 
 \subcaption{D$^4$M (IPC=50)}
    \label{mcp:3}
  \end{subfigure}
  \hfill
  \begin{subfigure}[b]{0.245\linewidth}
    \centering
    \includegraphics[width=1.0\linewidth]{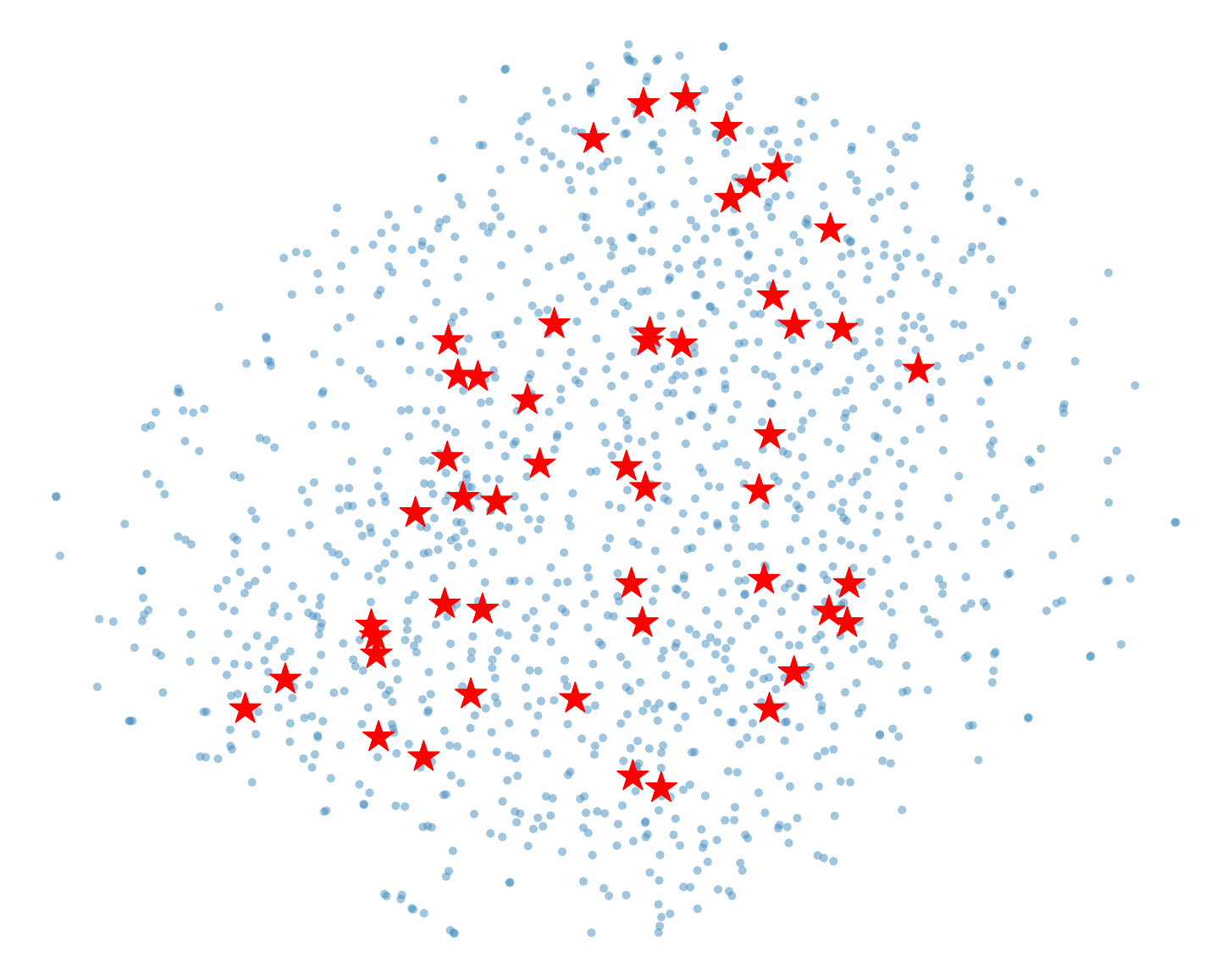} 
     \subcaption{DEO (IPC=50)}
    \label{mcp:4}
  \end{subfigure}
}
  \caption{
t-SNE visualization of latent distributions for original and distilled images. Results are shown for ImageWoof under IPC 10 and 50. Blue dots indicate original data; red stars denote distilled data from D$^4$M and DEO.}
 \vskip-15pt
  \label{vis_tsne}
\end{figure}
\begin{figure*}[t]
   \resizebox{0.87\linewidth}{!}{
  \includegraphics[width=1.0\textwidth]{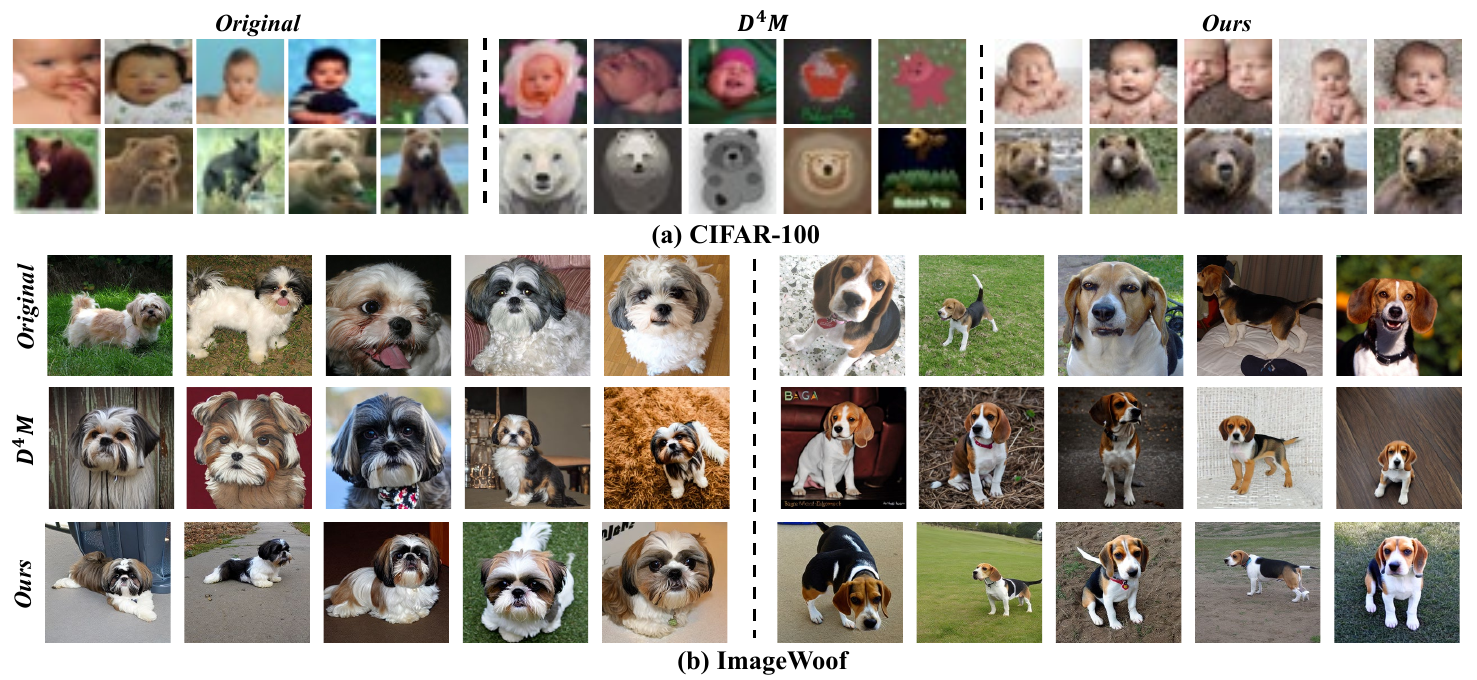}}
  \caption{Visualization of distilled data. We compare original samples with distilled data generated by D$^4$M and DEO on (a) CIFAR-100 (baby, bear) and (b) ImageWoof (Shih-Tzu, Beagle).
}
   \vspace{-0.4cm} 
  \label{vis_oursDM}
\end{figure*}

\textbf{Diversity Analysis.} The average cosine distance between samples and their class prototypes serves as a principled metric for intra-class diversity, which is fundamentally linked to the generalization capability of downstream models. To validate this, we report the average cosine distance for the first ten classes of CIFAR-100 and ImageNet-1K in Figure \ref{distance_imagewoof}. As illustrated, our DEO consistently outperforms the $\text{D}^4\text{M}$ baseline across both datasets. Specifically, on CIFAR-100, DEO achieves an average cosine distance of 0.143, marking a 9.2\% relative improvement over $\text{D}^4\text{M}$ (0.131). A similar trend is observed on ImageNet-1K, where DEO elevates the metric from 0.121 to 0.132 (a 9.1\% gain). These quantitative results demonstrate that DEO more effectively captures intrinsic semantic variability, thereby enriching the diversity of the distilled dataset and ultimately boosting downstream performance.

\textbf{Computational Efficiency.} We benchmark DEO against existing methods on a single NVIDIA A100 (80GB) GPU. As shown in Table~\ref{tab:comp_cost_vertical}, DEO substantially reduces the computational overhead of the distillation pipeline. On CIFAR-100, it decreases the per-class processing time from 9.5 to 7.4 minutes. The advantage is more pronounced on the large-scale ImageNet-1K dataset, where VLCP requires 25.2 minutes per class, mainly due to the heavy cost of LLaVA inference (21.7 minutes) and latent-space clustering (3.5 minutes). By removing these expensive stages and relying instead on efficient embedding-statistics computation with lightweight optimization, DEO reduces the total time to 13.4 minutes per class, yielding a 47\% reduction in computational cost while also improving Top-1 accuracy. These results demonstrate that DEO achieves better  performance with significantly lower resource demands.

\subsection{Visualization}

\textbf{Latent Space Visualization.} 
To qualitatively demonstrate the effectiveness of our DEO in capturing the intrinsic semantic structure of original data, we visualize the distilled embeddings of the same class on ImageWoof using t-SNE under IPC = 10 and 50, comparing D$^4$M with DEO as shown in Figure~\ref{vis_tsne}. In these plots, blue dots represent original samples, while red stars indicate distilled ones. For D$^4$M, the clustering-based approach results in distilled points that are relatively close to each other, with two points nearly overlapping under IPC=10, and even more clustered under IPC=50, suggesting limited diversity. In contrast, DEO produces embeddings that better align with the original data distribution and maintain a higher degree of dispersion in the latent space. This suggests that optimizing learnable embeddings enables the diffusion model to capture and reproduce meaningful semantic structures from the dataset, rather than memorizing superficial patterns.

\textbf{Distilled Images.} 
We further compare the visual quality and semantic fidelity of the distilled data generated by D$^4$M and our DEO on CIFAR-100 and ImageWoof in Figure~\ref{vis_oursDM}. Two major observations can be drawn: (1) On the low-resolution CIFAR-100, our DEO produces distilled samples that more faithfully retain the semantic characteristics of the original data. For example, in the bear and baby classes, DEO generates samples with diverse poses and expressions, whereas those from D$^4$M tend to look very similar.
(2) On the high-resolution ImageWoof, our DEO not only maintains clear class discriminability but also captures richer semantic diversity. For instance, in the Shih-Tzu class, D$^4$M frequently generates seated samples with similar hair color and style, whereas DEO produces samples in both seated and lying poses with diverse hair colors (e.g., brown and black), better capturing the true intra-class variability. These results show that DEO more effectively captures semantic structure, yielding distilled samples with higher realism and diversity.

\section{Conclusion}
In this paper, we introduce Disentangled Embedding Optimization (DEO), a novel clustering-free framework for diffusion-based dataset distillation. Instead of relying on latent clustering, DEO directly optimizes the learnable embeddings, which are utilized to generate distilled data, by matching the Gaussian statistics (class-level means and covariances) of the original data distribution. To further enhance the diversity among embeddings, a separation regularization term is incorporated to encourage inter-embedding decorrelation. Extensive experiments across multiple benchmarks demonstrate that the proposed DEO consistently outperforms prior methods. It generates more diversified and informative distilled data, demonstrates strong generalization across different model architectures, and maintains lower computational overhead.

\section{Acknowledgments}
This work was supported by Strategic Priority Research Program of the Chinese Academy of Sciences under grant No.XDB0690302.

\bibliographystyle{ACM-Reference-Format}
\bibliography{mm2026-2}

@inproceedings{lin2014microsoft,
    title        = {Microsoft coco: Common objects in context},
    author       = {Lin, Tsung-Yi and Maire, Michael and Belongie, Serge and Hays, James and Perona, Pietro and Ramanan, Deva and Doll{\'a}r, Piotr and Zitnick, C Lawrence},
    booktitle    = {Computer Vision--ECCV 2014: 13th European Conference, Zurich, Switzerland, September 6-12, 2014, Proceedings, Part V 13},
    pages        = {740--755},
    year         = {2014},
    organization = {Springer},
}

@article{kuznetsova2020open,
    title     = {The open images dataset v4: Unified image classification, object detection, and visual relationship detection at scale},
    author    = {Kuznetsova, Alina and Rom, Hassan and Alldrin, Neil and Uijlings, Jasper and Krasin, Ivan and Pont-Tuset, Jordi and Kamali, Shahab and Popov, Stefan and Malloci, Matteo and Kolesnikov, Alexander and others},
    journal   = {International journal of computer vision},
    volume    = {128},
    number    = {7},
    pages     = {1956--1981},
    year      = {2020},
    publisher = {Springer},
}

@inproceedings{10.1145/3503161.3549202,
    author    = {Wong, Yongkang and Fan, Shaojing and Guo, Yangyang and Xu, Ziwei and Stephen, Karen and Sheoran, Rishabh and Bhamidipati, Anusha and Barsopia, Vivek and Liu, Jianquan and Kankanhalli, Mohan},
    title     = {Compute to Tell the Tale: Goal-Driven Narrative Generation},
    year      = {2022},
    isbn      = {9781450392037},
    publisher = {Association for Computing Machinery},
    address   = {New York, NY, USA},
    url       = {https://doi.org/10.1145/3503161.3549202},
    doi       = {10.1145/3503161.3549202},
    booktitle = {Proceedings of the 30th ACM International Conference on Multimedia},
    pages     = {6875–6882},
    numpages  = {8},
    series    = {MM '22},
}

@inproceedings{10.1145/3774904.3792350,
    author    = {Guan, Zenghao and Zhu, Guojun and Zhou, Yucan and Liu, Wu and Wang, Weiping and Luo, Jiebo and Gu, Xiaoyan},
    title     = {Enhancing Federated Class-Incremental Learning via Spatial-Temporal Statistics Aggregation},
    year      = {2026},
    isbn      = {9798400723070},
    publisher = {Association for Computing Machinery},
    address   = {New York, NY, USA},
    url       = {https://doi.org/10.1145/3774904.3792350},
    doi       = {10.1145/3774904.3792350},
    booktitle = {Proceedings of the ACM Web Conference 2026},
    pages     = {5356–5367},
    numpages  = {12},
    location  = {United Arab Emirates},
    series    = {WWW '26},
}

@ARTICLE{11329188,
    author   = {Liu, Yijun and Liu, Wu and Gu, Xiaoyan and Rui, Yong and He, Xiaodong and Zhang, Yongdong},
    journal  = {IEEE Transactions on Multimedia},
    title    = {LMAgent: A Large-scale Multimodal Agents Society for Multi-user Simulation},
    year     = {2026},
    pages    = {1-12},
    doi      = {10.1109/TMM.2026.3651078},
}

@article{wang2018dataset,
    title   = {Dataset distillation},
    author  = {Wang, Tongzhou and Zhu, Jun-Yan and Torralba, Antonio and Efros, Alexei A},
    journal = {arXiv preprint arXiv:1811.10959},
    year    = {2018},
}

@article{yu2023review,
    title     = {A Comprehensive Survey to Dataset Distillation},
    author    = {Yu, Ruonan and Liu, Songhua and Wang, Xinchao},
    journal   = {IEEE Transactions on Pattern Analysis and Machine Intelligence},
    volume    = {46},
    number    = {1},
    pages     = {150--170},
    year      = {2023},
    publisher = {IEEE},
}

@article{lei2023survey,
    title     = {A Comprehensive Survey to Dataset Distillation},
    author    = {Lei, Shiye and Tao, Dacheng},
    journal   = {IEEE Transactions on Pattern Analysis and Machine Intelligence},
    volume    = {46},
    number    = {1},
    pages     = {17--32},
    year      = {2023},
    publisher = {IEEE},
}

@inproceedings{dong2022privacy,
    title     = {Privacy for Free: How does Dataset Condensation Help Privacy?},
    author    = {Dong, Tian and Zhao, Bo and Liu, Lingjuan},
    booktitle = {Proceedings of the International Conference on Machine Learning (ICML)},
    pages     = {5378--5396},
    year      = {2022},
}

@inproceedings{zheng2025dosser,
    title     = {Improving Noise Efficiency in Privacy-preserving Dataset Distillation},
    author    = {Zheng, Runkai and Dasu, Vishnu Asutosh and Wang, Yinong Oliver and Wang, Haohan and Torre, Fernando De la},
    booktitle = {Proceedings of the IEEE/CVF International Conference on Computer Vision (ICCV)},
    year      = {2025},
}

@inproceedings{yang2023efficient,
    title     = {An Efficient Dataset Condensation Plugin and Its Application to Continual Learning},
    author    = {Yang, Enneng and Shen, Li and Wang, Zhenyi and Liu, Tongliang and Guo, Guibing},
    booktitle = {Proceedings of the Advances in Neural Information Processing Systems (NeurIPS)},
    year      = {2023},
}

@inproceedings{gu2024ssd,
    title     = {Summarizing Stream Data for Memory-Restricted Online Continual Learning},
    author    = {Gu, Jianyang and Wang, Kai and Jiang, Wei and You, Yang},
    booktitle = {Proceedings of the AAAI Conference on Artificial Intelligence (AAAI)},
    year      = {2024},
}

@inproceedings{cazenavette2022dataset,
    title     = {Dataset Distillation by Matching Training Trajectories},
    author    = {Cazenavette, George and Wang, Tongzhou and Torralba, Antonio and Efros, Alexei A. and Zhu, Jun-Yan},
    booktitle = {Proceedings of the IEEE/CVF Conference on Computer Vision and Pattern Recognition (CVPR)},
    pages     = {10718--10727},
    year      = {2022},
}

@inproceedings{du2023minimizing,
    title     = {Minimizing the accumulated trajectory error to improve dataset distillation},
    author    = {Du, Jiawei and Jiang, Yidi and Tan, Vincent YF and Zhou, Joey Tianyi and Li, Haizhou},
    booktitle = {Proceedings of the IEEE/CVF conference on computer vision and pattern recognition},
    pages     = {3749--3758},
    year      = {2023},
}

@inproceedings{liu2024att,
    title     = {Dataset Distillation by Automatic Training Trajectories},
    author    = {Liu, Dai and Gu, Jindong and Cao, Hu and Trinitis, Carsten and Schulz, Martin},
    booktitle = {Proceedings of the European Conference on Computer Vision (ECCV)},
    year      = {2024},
}

@inproceedings{zhao2021datasetcondensation,
    title     = {Dataset Condensation with Gradient Matching},
    author    = {Zhao, Bo and Bilen, Hakan},
    booktitle = {Proceedings of the International Conference on Learning Representations (ICLR)},
    year      = {2021},
}

@inproceedings{zhao2021differentiatble,
    title     = {Dataset condensation with Differentiable Siamese Augmentation},
    author    = {Zhao, Bo and Bilen, Hakan},
    booktitle = {Proceedings of the International Conference on Machine Learning (ICML)},
    pages     = {12674--12685},
    year      = {2021},
}

@inproceedings{lee2022dataset,
    title        = {Dataset condensation with contrastive signals},
    author       = {Lee, Saehyung and Chun, Sanghyuk and Jung, Sangwon and Yun, Sangdoo and Yoon, Sungroh},
    booktitle    = {International Conference on Machine Learning},
    pages        = {12352--12364},
    year         = {2022},
    organization = {PMLR},
}

@inproceedings{zhao2023distribution,
    title     = {Dataset Condensation with Distribution Matching},
    author    = {Zhao, Bo and Bilen, Hakan},
    booktitle = {Proceedings of the IEEE/CVF Winter Conference on Applications of Computer Vision (WACV)},
    pages     = {6514--6523},
    year      = {2023},
}

@inproceedings{wang2022cafe,
    title     = {{CAFE}: Learning to Condense Dataset by Aligning Features},
    author    = {Wang, Kai and Zhao, Bo and Peng, Xiangyu and Zhu, Zheng and Yang, Shuo and Wang, Shuo and Huang, Guan and Bilen, Hakan and Wang, Xinchao and You, Yang},
    booktitle = {Proceedings of the IEEE/CVF Conference on Computer Vision and Pattern Recognition (CVPR)},
    pages     = {12196--12205},
    year      = {2022},
}

@inproceedings{zhao2023improved,
    title     = {Improved distribution matching for dataset condensation},
    author    = {Zhao, Ganlong and Li, Guanbin and Qin, Yipeng and Yu, Yizhou},
    booktitle = {Proceedings of the IEEE/CVF Conference on Computer Vision and Pattern Recognition},
    pages     = {7856--7865},
    year      = {2023},
}

@inproceedings{yin2023squeeze,
    title     = {Squeeze, Recover and Relabel: Dataset Condensation at ImageNet Scale From A New Perspective},
    author    = {Yin, Zeyuan and Xing, Eric and Shen, Zhiqiang},
    booktitle = {Thirty-seventh Conference on Neural Information Processing Systems},
    year      = {2023},
}

@inproceedings{zhao2025d3hr,
    title     = {Taming Diffusion for Dataset Distillation with High Representativeness},
    author    = {Zhao, Lin and Wu, Yushu and Jiang, Xinru and Gu, Jianyang and Wang, Yanzhi and Xu, Xiaolin and Zhao, Pu and Lin, Xue},
    booktitle = {Proceedings of the International Conference on Machine Learning (ICML)},
    year      = {2025},
}

@inproceedings{chen2025igd,
    title     = {Influence-Guided Diffusion for Dataset Distillation},
    author    = {Chen, Mingyang and Du, Jiawei and Huang, Bo and Wang, Yi and Zhang, Xiaobo and Wang, Wei},
    booktitle = {Proceedings of the International Conference on Learning Representations (ICLR)},
    year      = {2025},
}

@article{liu2025survey,
    title   = {The Evolution of Dataset Distillation: Toward Scalable and Generalizable Solutions},
    author  = {Liu, Ping and Du, Jiawei},
    journal = {arXiv preprint arXiv:2502.05673},
    year    = {2025},
}

@inproceedings{cazenavette2023generalizing,
    title     = {Generalizing dataset distillation via deep generative prior},
    author    = {Cazenavette, George and Wang, Tongzhou and Torralba, Antonio and Efros, Alexei A and Zhu, Jun-Yan},
    booktitle = {Proceedings of the IEEE/CVF conference on computer vision and pattern recognition},
    pages     = {3739--3748},
    year      = {2023},
}

@InProceedings{Su_2024_CVPR,
    author    = {Su, Duo and Hou, Junjie and Gao, Weizhi and Tian, Yingjie and Tang, Bowen},
    title     = {D{\textasciicircum}4M: Dataset Distillation via Disentangled Diffusion Model},
    booktitle = {Proceedings of the IEEE/CVF Conference on Computer Vision and Pattern Recognition (CVPR)},
    month     = {June},
    year      = {2024},
    pages     = {5809-5818},
}

@inproceedings{zou2025vlcp,
    title     = {Dataset Distillation via Vision-Language Category Prototype},
    author    = {Zou, Yawen and Li, Guang and Su, Duo and Wang, Zi and Yu, Jun and Zhang, Chao},
    booktitle = {Proceedings of the IEEE/CVF International Conference on Computer Vision (ICCV)},
    year      = {2025},
}

@inproceedings{kim2022dataset,
    title        = {Dataset condensation via efficient synthetic-data parameterization},
    author       = {Kim, Jang-Hyun and Kim, Jinuk and Oh, Seong Joon and Yun, Sangdoo and Song, Hwanjun and Jeong, Joonhyun and Ha, Jung-Woo and Song, Hyun Oh},
    booktitle    = {International Conference on Machine Learning},
    pages        = {11102--11118},
    year         = {2022},
    organization = {PMLR},
}

@inproceedings{cui2023scaling,
    title        = {Scaling up dataset distillation to imagenet-1k with constant memory},
    author       = {Cui, Justin and Wang, Ruochen and Si, Si and Hsieh, Cho-Jui},
    booktitle    = {International Conference on Machine Learning},
    pages        = {6565--6590},
    year         = {2023},
    organization = {PMLR},
}

@inproceedings{sajedi2023datadam,
    title     = {Datadam: Efficient dataset distillation with attention matching},
    author    = {Sajedi, Ahmad and Khaki, Samir and Amjadian, Ehsan and Liu, Lucy Z and Lawryshyn, Yuri A and Plataniotis, Konstantinos N},
    booktitle = {Proceedings of the IEEE/CVF International Conference on Computer Vision},
    pages     = {17097--17107},
    year      = {2023},
}

@inproceedings{zhang2024m3d,
    title     = {M3D: Dataset Condensation by Minimizing Maximum Mean Discrepancy},
    author    = {Hansong Zhang and Shikun Li and Pengju Wang and Dan Zeng and Shiming Ge},
    year      = {2024},
    booktitle = {The 38th Annual AAAI Conference on Artificial Intelligence (AAAI)},
}

@inproceedings{zhang2024dance,
    title     = {{DANCE}: Dual-View Distribution Alignment for Dataset Condensation},
    author    = {Zhang, Hansong and Li, Shikun and Lin, Fanzhao and Wang, Weiping and Qian, Zhenxing and Ge, Shiming},
    booktitle = {Proceedings of the International Joint Conference on Artificial Intelligence (IJCAI)},
    year      = {2024},
}

@inproceedings{deng2024iid,
    title     = {Exploiting Inter-sample and Inter-feature Relations in Dataset Distillation},
    author    = {Deng, Wenxiao and Li, Wenbin and Ding, Tianyu and Wang, Lei and Zhang, Hongguang and Huang, Kuihua and Huo, Jing and Gao, Yang},
    booktitle = {Proceedings of the IEEE/CVF Conference on Computer Vision and Pattern Recognition (CVPR)},
    pages     = {17057--17066},
    year      = {2024},
}

@inproceedings{li2024dsdm,
    title     = {Diversified Semantic Distribution Matching for Dataset Distillation},
    author    = {Li, Hongcheng and Zhou, Yucan and Gu, Xiaoyan and Li, Bo and Wang, Weiping},
    booktitle = {Proceedings of the ACM International Conference on Multimedia (MM)},
    year      = {2024},
}

@article{yin2023dataset,
    title   = {Dataset Distillation via Curriculum Data Synthesis in Large Data Era},
    author  = {Yin, Zeyuan and Shen, Zhiqiang},
    journal = {Transactions on Machine Learning Research},
    year    = {2024},
}

@InProceedings{sun2024diversity,
    title     = {On the Diversity and Realism of Distilled Dataset: An Efficient Dataset Distillation Paradigm},
    author    = {Sun, Peng and Shi, Bei and Yu, Daiwei and Lin, Tao},
    booktitle = {Proceedings of the IEEE/CVF Conference on Computer Vision and Pattern Recognition (CVPR)},
    year      = {2024},
}

@inproceedings{shao2024gvbsm,
    title     = {Generalized Large-Scale Data Condensation via Various Backbone and Statistical Matching},
    author    = {Shao, Shitong and Yin, Zeyuan and Zhou, Muxin and Zhang, Xindong and Shen, Zhiqiang},
    booktitle = {Proceedings of the IEEE/CVF Conference on Computer Vision and Pattern Recognition (CVPR)},
    pages     = {16709--16718},
    year      = {2024},
}

@inproceedings{dwa2024neurips,
    title     = {Diversity-Driven Synthesis: Enhancing Dataset Distillation through Directed Weight Adjustment},
    author    = {Du, Jiawei and Zhang, Xin and Hu, Juncheng and Huang, Wenxin and Zhou, Joey Tianyi},
    booktitle = {Adv. Neural Inf. Process. Syst. (NeurIPS)},
    year      = {2024},
}

@inproceedings{li2025deda,
    title     = {Diversity-Enhanced Distribution Alignment for Dataset Distillation},
    author    = {Li, Hongcheng and Zhou, Yucan and Gu, Xiaoyan and Li, Bo and Wang, Weiping},
    booktitle = {Proceedings of the IEEE/CVF International Conference on Computer Vision (ICCV)},
    year      = {2025},
}

@inproceedings{liu2025autoprompt,
    title     = {AutoPrompt: Automated Red-Teaming of Text-to-Image Models via LLM-Driven Adversarial Prompts},
    author    = {Liu, Yufan and Zhang, Wanqian and Chen, Huashan and Wang, Lin and Jia, Xiaojun and Lin, Zheng and Wang, Weiping},
    booktitle = {Proceedings of the IEEE/CVF International Conference on Computer Vision},
    pages     = {17557--17566},
    year      = {2025},
}

@inproceedings{gu2024efficient,
    title     = {Efficient dataset distillation via minimax diffusion},
    author    = {Gu, Jianyang and Vahidian, Saeed and Kungurtsev, Vyacheslav and Wang, Haonan and Jiang, Wei and You, Yang and Chen, Yiran},
    booktitle = {Proceedings of the IEEE/CVF Conference on Computer Vision and Pattern Recognition},
    pages     = {15793--15803},
    year      = {2024},
}

@inproceedings{santiagomgd,
  title     = {{MGD}$^3$: Mode-Guided Dataset Distillation using Diffusion Models},
  author    = {Chan Santiago, Jeffrey A. and Tirupattur, Praveen and Nayak, Gaurav Kumar and Liu, Gaowen and Shah, Mubarak},
  booktitle = {Proceedings of the 42nd International Conference on Machine Learning (ICML)},
  year      = {2025},
}

@article{krizhevsky2009learning,
    title     = {Learning multiple layers of features from tiny images},
    author    = {Krizhevsky, Alex and Hinton, Geoffrey and others},
    year      = {2009},
    publisher = {Toronto, ON, Canada},
}

@inproceedings{deng2009imagenet,
    title        = {Imagenet: A large-scale hierarchical image database},
    author       = {Deng, Jia and Dong, Wei and Socher, Richard and Li, Li-Jia and Li, Kai and Fei-Fei, Li},
    booktitle    = {2009 IEEE conference on computer vision and pattern recognition},
    pages        = {248--255},
    year         = {2009},
    organization = {Ieee},
}

@misc{imagenette_howard_2019,
    author       = {Howard, Jeremy},
    title        = {Imagenette: A smaller subset of 10 easily classified classes from {ImageNet} (and a little more French)},
    howpublished = {\url{https://github.com/fastai/imagenette}},
    note         = {GitHub repository},
    year         = {2019},
}

@inproceedings{sener2018active,
    title     = {Active Learning for Convolutional Neural Networks: A Core-Set Approach},
    author    = {Sener, Ozan and Savarese, Silvio},
    booktitle = {International Conference on Learning Representations},
    year      = {2018},
}

@inproceedings{10.1145/1553374.1553517,
    author    = {Welling, Max},
    title     = {Herding dynamical weights to learn},
    year      = {2009},
    isbn      = {9781605585161},
    publisher = {Association for Computing Machinery},
    address   = {New York, NY, USA},
    url       = {https://doi.org/10.1145/1553374.1553517},
    doi       = {10.1145/1553374.1553517},
    booktitle = {Proceedings of the 26th Annual International Conference on Machine Learning},
    pages     = {1121–1128},
    numpages  = {8},
    location  = {Montreal, Quebec, Canada},
    series    = {ICML '09},
}

@inproceedings{peebles2023scalable,
    title     = {Scalable diffusion models with transformers},
    author    = {Peebles, William and Xie, Saining},
    booktitle = {Proceedings of the IEEE/CVF international conference on computer vision},
    pages     = {4195--4205},
    year      = {2023},
}

\end{document}